\RequirePackage{fix-cm}
\makeatletter
\def\cl@chapter{}
\makeatother
\documentclass[twocolumn]{svjour3}          %
\usepackage[switch]{lineno}
\smartqed  
\usepackage{graphicx}
\usepackage{algorithm}
\usepackage[noend]{algorithmic}
\usepackage{amsmath}
\usepackage{amssymb}
\usepackage{booktabs}
\usepackage{svg}
\usepackage{hyperref} 

\hypersetup{
    colorlinks=false,
    hidelinks
}
\usepackage[capitalize]{cleveref}
\usepackage{multirow}
\usepackage{stackengine}
\usepackage{mathtools}
\usepackage{wrapfig}
\usepackage{lipsum}
\usepackage{pifont}
\usepackage{url} 
\usepackage{placeins}
\usepackage{ulem}
\renewcommand{\v}[1]{\boldsymbol{#1}} 

\usepackage{ifthen}

\newboolean{showcomments}
\setboolean{showcomments}{false}  

\ifthenelse{\boolean{showcomments}}{
  \usepackage[colorinlistoftodos,prependcaption,textsize=tiny]{todonotes}
  \newcommand{\rone}[2][1=]{\todo[linecolor=blue,backgroundcolor=blue!25,bordercolor=blue]{\textbf{#2}}}
  \newcommand{\rtwo}[2][1=]{\todo[linecolor=orange,backgroundcolor=orange!25,bordercolor=orange]{\textbf{#2}}}
  \newcommand{\rthree}[2][1=]{\todo[linecolor=olive,backgroundcolor=olive!25,bordercolor=olive]{\textbf{#2}}}
  \newcommand{\changeone}[1]{{\textcolor{blue} {#1}}}

}{
  \newcommand{\rone}[2]{}
  \newcommand{\rtwo}[2]{}
  \newcommand{\rthree}[2]{}
  \newcommand{\changeone}[1]{#1}

}

\newcommand{\where}[1]{\textcolor{blue}{{#1}}}
\begin{document}
\begin{sloppypar}
\title{HMI: Hierarchical Knowledge Management for Efficient Multi-Tenant Inference in Pretrained Language Models}

\author{Jun Zhang \and Jue Wang \and Huan Li \and Lidan Shou \and Ke Chen \and Gang Chen \and Qin Xie \and Guiming Xie \and Xuejian Gong}

\institute{Jun Zhang \and Jue Wang \and Huan Li \and Lidan Shou \and Ke Chen \and Gang Chen \at
              1. The State Key Laboratory of Blockchain and Data Security, Zhejiang University \\
              2. Hangzhou High-Tech Zone (Binjiang) Institute of Blockchain and Data Security \\
              \email{\{zj.cs,zjuwangjue,lihuan.cs,should,chenk,cg\}@zju.edu.cn}
           \and
           Qin Xie \and Guiming Xie \and Xuejian Gong \at
              OPPO AI Center\\
            \email{\{xieq,xieguiming,gongxuejian\}@oppo.com}
}

\date{Received: date / Accepted: date}

\maketitle

\begin{abstract}
The significant computational demands of pretrained language models (PLMs), which often require dedicated hardware, present a substantial challenge in serving them efficiently, especially in multi-tenant environments. To address this, we introduce HMI, a Hierarchical knowledge management-based Multi-tenant Inference system, designed to manage tenants with distinct PLMs resource-efficiently. Our approach is three-fold: Firstly, we categorize PLM knowledge into general, domain-specific, and task-specific. Leveraging insights on knowledge acquisition across different model layers, we construct hierarchical PLMs (\textsf{hPLM}s) by extracting and storing knowledge at different levels, significantly reducing GPU memory usage per tenant. Secondly, we establish hierarchical knowledge management for \textsf{hPLM}s generated by various tenants in HMI. We manage domain-specific knowledge with acceptable storage increases by constructing and updating domain-specific knowledge trees based on frequency. We manage task-specific knowledge within limited GPU memory through parameter swapping. Finally, we propose system optimizations to enhance resource utilization and inference throughput. These include fine-grained pipelining via hierarchical knowledge prefetching to overlap CPU and I/O operations with GPU computations, and optimizing parallel implementations with batched matrix multiplications.
Our experimental results demonstrate that the proposed HMI can efficiently serve up to 10,000 \textsf{hPLM}s (\textsf{hBERT}s and \textsf{hGPT}s) on a single GPU, with only a negligible compromise in accuracy.
\end{abstract}

\section{Introduction}
Deep learning is being incorporated into a growing number of cloud services \cite{chard2019dlhub,li2020automating}.
Recently, a series of deep learning models known as pre-trained language models (PLMs)
\cite{bert,gpt,roberta,brown2020language} have become prevalent in the field of natural language processing (NLP).
These models, pre-trained on broad data, can learn universal language representations, and thereby achieve exceptional results on a wide range of downstream NLP tasks. 
It leads to a dominant trend emerging in the AI landscape, where a single PLM is utilized across diverse tasks, referred to as `homogenization' \cite{bommasani2021opportunities}. 
This trend aligns seamlessly with the principles of cloud computing, 
as cloud platforms can efficiently train and deploy PLMs as part of their core infrastructure, and amortize these costs by serving massive numbers of tenants. 

\begin{figure}[!ht]
    \centering
    \includegraphics[width=1\linewidth]{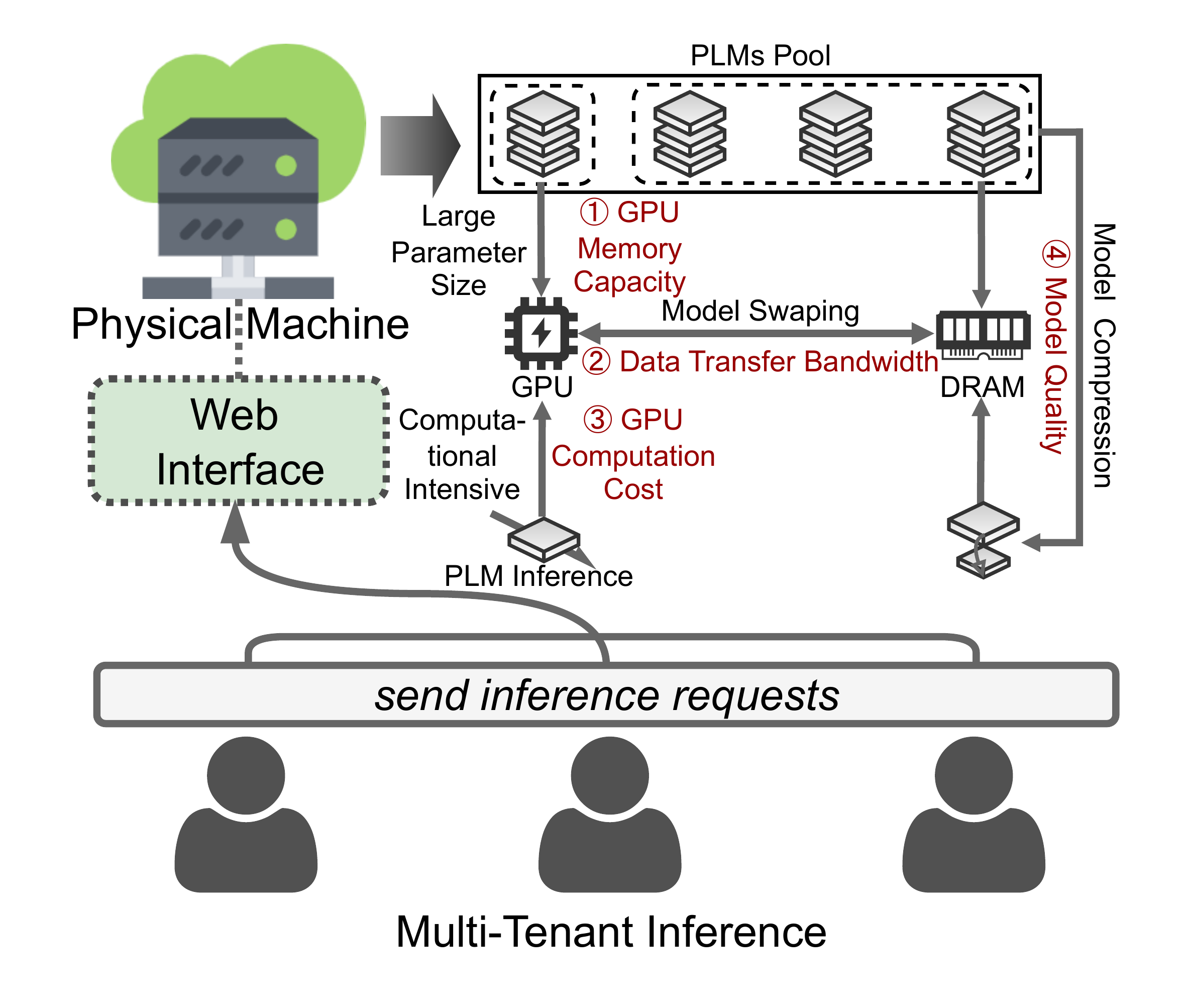}
    \caption{
    An overview of the multi-tenant inference workflow and the associated challenges when adapted for PLMs.}
    \label{fig:sys_com}
\end{figure}

\begin{figure*}[!t]
    \centering
    \includegraphics[width=0.8\linewidth]{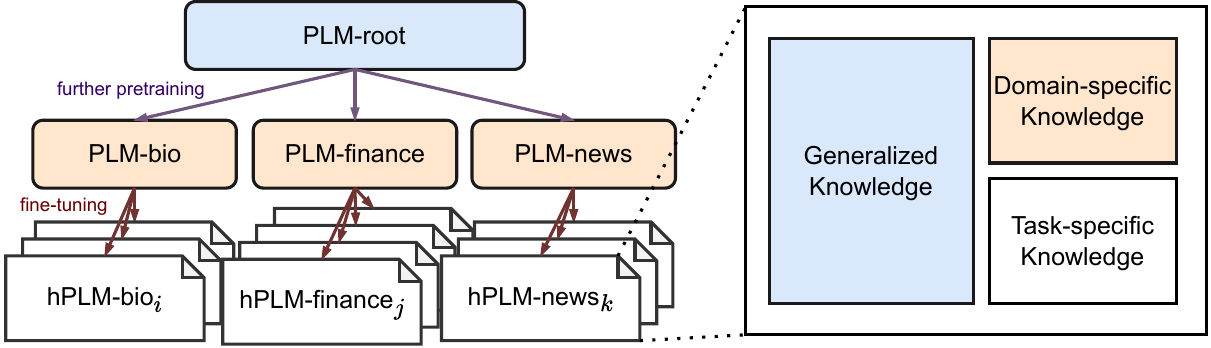}
    \caption{Model version tree and the separation of the learned hierarchical knowledge of a \textsf{hPLM} instance.}
    \label{fig:branch}
\end{figure*}

However, one single
generic PLM cannot be the silver bullet for all scenarios, 
as it may not perform well for \textit{domain-specific} or \textit{highly customized} tasks.
For example, an online grocery retailing platform may experience low accuracy with a generic PLM, and thus require further-pretraining \cite{jin2021lifelong} it on different retail domains.
Moreover, each store (tenant) of the platform, selling its own product portfolio, may need to additionally fine-tune the model \cite{howard2018universal} over its proprietary data for per-store services such as chat-bots and feedback data analysis. 
The above further-pretraining and fine-tuning process can result in hundreds or even thousands of distinct model copies from a generic PLM.
While training can be conducted during off-peak hours, with training costs amortized over the number of tenants and the duration of their subscriptions, provisioning inference services for distinct, customized PLMs in the cloud remains a significant challenge.
\textit{Multi-tenant inference}, a general paradigm that co-locates multiple models on the same physical machine, has been widely adopted to improve hardware utilization, increase throughput, and reduce energy costs~\cite{baek2020multi,liu2022veltair}. 
However, due to the characteristics of PLMs, including large parameter sizes, substantial memory demands, and intensive computational requirements, extending this paradigm introduces new challenges in achieving efficient multi-tenant inference for PLMs.
For instance, as illustrated in~\cref{fig:sys_com}, the cloud platform operator utilizes a physical machine hosting 12 distinct PLMs, each handling tenant-specific inference requests through a shared web interface:

\textbf{Challenge 1 (GPU Memory Capacity).}
Accommodating a massive number of models on limited hardware is non-trivial, as even a moderately sized model, e.g.,~a customized BERT \cite{bert}, requires several hundred MBs of GPU memory. This means that a typical GPU, e.g.,~NVIDIA Quadro P5000 with 16GB VRAM, can only host a few models at most.
As a workaround, models can be stored in the host memory or SSDs and swapped into the GPU memory for computation only when an \texttt{infer} request arrives.
For example, due to the limited GPU capacity of \textcircled{1} in~\cref{fig:sys_com}, 3 PLMs from the PLMs Pool can be hosted on the GPU, while the remaining 9 PLMs need to be stored in DRAM (including host memory and SSDs).
However, model swapping introduces the subsequent challenge.

\textbf{Challenge 2 (Data Transfer Bandwidth).} 
Since PLMs are large in model size, the I/O overhead of model swapping becomes the bottleneck and significantly reduces system efficiency.
As shown in~\textcircled{2} of~\cref{fig:sys_com}, with constrained data transfer bandwidth, frequent swapping (loading and offloading) of 9 PLMs between DRAM and GPU memory results in increased inference latency and reduced throughput.

\textbf{Challenge 3 (GPU Computation Cost).}  
Additionally, as PLMs are computationally intensive and incur high GPU computation cost (see \cref{fig:sys_com} \textcircled{3}), running multiple models on a single GPU leads to competition for computing resources, resulting in low throughput and high latency.

\textbf{Challenge 4 (Model Quality).}
We also note that conventional \textit{lossy} model compression methods, e.g., distillation \cite{hinton2015distilling}, quantization \cite{han2015deep}, pruning \cite{reed1993pruning}, might not come to the rescue in this circumstance, as they may compromise the accuracy and fail to meet the service-level requirement.
More importantly, despite reducing PLM size at the expense of model quality (see \cref{fig:sys_com} \textcircled{4}), serving thousands of independent (compressed) PLMs on a single GPU remains challenging, and thus, model swapping is still required, leaving Challenge 2 unresolved.

In this paper, we investigate whether massive customized model instances can be efficiently served on a single GPU while maintaining throughput and model quality. 
We demonstrate that our proposed multi-tenant inference system achieves this by hierarchically extracting, storing, and managing knowledge in PLMs from different tenants.
Each tenant uses a standalone hierarchical PLM (\textsf{hPLM}) without interference from others, while all tenants' \textsf{hPLM}s share the physical hardware.
For the \textsf{hPLM} type, we show the effectiveness of the system in serving a large number of hBERT (encoder-only PLM) and hGPT (decoder-only PLM) on a single GPU in \cref{sec:evaluation}.

Like a typical PLM, a \textsf{hPLM} needs to go through the following three training phases before being ready to serve \texttt{infer} requests: 
1) pretraining (PT) on a large, domain-agnostic corpus to acquire generalized knowledge, 
2) further pretraining  (FPT) on a domain-specific dataset to gain domain-specific knowledge, 
and 3) fine-tuning (FT) on task-specific data to additionally obtain task-specific knowledge. 

It is important to note that, while all models are identical after pretraining, their neural weights are updated differently after FPT and FT. 
Taking BERT as an example, all models to be provisioned comprise a logical version tree structure as depicted in \cref{fig:branch}, where the root node is a generic BERT from PT, the internal nodes represent models from FPT, and the leaf nodes represent those from FT. 
The edges of the tree indicate weight updates made during FPT or FT.
In practice, the cloud platform performs PT and FPT to construct the root and branch versions of BERT. 
Tenants choose one of these versions and upload their own data for FT to create custom \textsf{hPLM} instances.

In this paper, we present a co-design of cost-effective storage and computation scheme for managing a large number of \textsf{hPLM} instances. 
Starting from an interesting observation, we found that FPT and FT learn domain-specific and task-specific knowledge at different layers of the model, respectively. Specifically, by dividing the PLM into two parts, domain-specific knowledge is captured in the lower layers (roughly the first half) during FPT, while task-specific knowledge is captured in the higher layers (roughly the second half) during FT.
The representational properties of deep neural networks have been extensively studied, as discussed in~\cite{stephenson2021on}.
Therefore, aside from the general 
knowledge learned of the root model during PT in the version tree, during FPT, we freeze the higher layers of the PLM and update only the lower layers to extract domain-specific knowledge. Conversely, we freeze the lower layers and update only the higher layers to extract task-specific knowledge during FT.
Thus, instead of storing the models, we propose to extract the \textit{knowledge} learned by the leaf-node models (\textsf{hPLM}s) of the version tree and store this knowledge in two compact data structures. 
The first data structure is a \textit{precomputed lookup table} (PLOT) that captures generic and domain-specific knowledge \cite{wang2022skipbert}. 
It is essentially a key-value table with text fragments as keys and their corresponding internal representations of \textsf{hPLM} as values. 
PLOT is designed to approximate the internal representation of any input text by performing a series of lookup operations and an aggregation. 
Since PLOT replaces GPU computation with CPU lookup operations, it does not use any additional GPU memory for each \textsf{hPLM}.
Second, the task-specific knowledge is captured in a set of compact neural blocks called 
\textit{adapters} \cite{houlsby2019parameter}~\footnote{In this paper, we use ``FT" as a unified term to refer to general task-specific performance improvements through fine-tuning. The term ``adapter" specifically refers to the FT method used for extracting and storing task-specific knowledge in the HMI.}. 
Adapters can be embedded into PLM (root or internal nodes) to obtain a task-specific \textsf{hPLM} (leaf of the version tree of \cref{fig:branch}). 
In our system, we maintain shared high layers of the PLM transformer (from PLM-root). When tenants submit inference tasks involving multiple different \textsf{hPLM}s, we first obtain the internal representations of the \textsf{hPLM} lower layers through PLOT, and then feed these representations into the \textsf{hPLM} high layers, which combine task-specific adapters with the shared transformer layers, to compute the final results.

However, domain-specific PLOT introduces additional storage requirements. Given our finding that the difference in precomputed text representations across domains is minimal, we propose constructing a PLOT version tree and employing a frequency-based strategy to update only the most frequently used text fragments. This approach allows us to manage PLOTs for multiple domains with an acceptable increase in storage.
Additionally, for \textsf{hPLM}s, PLOT introduces lookup operations on the CPU. During batch inference, the system addresses this by prefetching domain-specific knowledge, thereby overlapping the CPU operations with GPU operations to further hide the lookup overhead.
For adapters, despite their compact form encapsulating task-specific knowledge, maintaining hundreds or thousands of them on a single GPU is impractical. To manage this task-specific knowledge within limited GPU memory, the system employs a swapping strategy: adapters are kept in the host memory and loaded to the GPU on-demand during the respective \textsf{hPLM} \texttt{infer} requests. To mitigate the additional I/O costs from adapter swapping, the system prefetches task-specific knowledge layer-wise, overlapping I/O operations with GPU computations of the transformer layer. Additionally, we utilize batched matrix multiplication to enable parallel computation of adapters for different requests within the same batch at the layer level, further improving inference efficiency.

Therefore, the comprehensive scheme of hierarchical knowledge construction, management, and prefetching consumes trivial additional memory in GPU for each \textsf{hPLM}, 
making it possible for a single-GPU machine to serve thousands of \textsf{hPLM} instances. 
To summarize, our key contributions are as follows:
\begin{itemize}
\item \textit{Novel System Design:} 
We propose a cost-effective Hierarchical knowledge management-based Multi-tenant Inference framework, called HMI, to accommodate thousands of custom hierarchical PLM (\textsf{hPLM}) instances on a single commodity cloud server with one GPU.

\item \textit{Hierarchical PLM Construction:}
We propose to construct a hierarchical PLM (\textsf{hPLM}) for each tenant by hierarchically extracting and storing model knowledge as general, domain-specific, and task-specific knowledge. For \textsf{hPLM} instances, in addition to the general knowledge acquired during PT, interesting findings suggest that FPT and FT learn domain-specific and task-specific knowledge at different transformer layers. Therefore, (a) domain-specific knowledge is captured and materialized as hidden states of the lower model layers into PLOT; and (b) task-specific knowledge is encapsulated in the adapter parameters of the higher model layers.
    
\item \textit{Hierarchical Knowledge Management:}
    Faced with customized \textsf{hPLM}s from numerous tenants, we introduce a frequency-based updating mechanism for domain-specific knowledge trees and a task-specific knowledge swapping strategy. These techniques enable HMI to be effectively managed within acceptable storage increases and limited GPU memory.

\item \textit{Hierarchical Knowledge Prefetching:} 
    We show that HMI can be optimized with hierarchical knowledge prefetching.
    This strategy hides additional overhead of CPU and I/O operations in the GPU operation, 
    further improving the overall system performance by 25\%.

\item \textit{Extensive Evaluation:} 
    Experimental results demonstrate that our system can simultaneously serve up to 10,000 \textsf{hPLM}s with one GPU without reducing the inference throughput. 
    We also extend our scheme to generative language models and achieve similar results.
\end{itemize}


The rest of the paper is organized as follows: 
\cref{sec:preli} describes the motivations of this paper.
\cref{sec:vbert} shows the construction of a single hierarchical PLM (\textsf{hPLM}) and the corresponding inference procedure of \textsf{hPLM}.
\cref{sec:system} illustrates the overall system architecture of HMI and the two types of requests it handles. 
\cref{sec:hkm} explains the hierarchical knowledge management approach of HMI. 
\cref{sec:optimization} details the system implementation and optimization. 
\cref{sec:evaluation} reports the experimental results. 
\cref{sec:related} discusses the related work.
\cref{sec:conclusion} concludes the paper. 
A preliminary demonstration of our work was presented at SIGMOD~\cite{wang2023smile}, where we introduced a prototype for handling massive PLMs in the cloud. 
Compared to the demonstration, this paper introduces the novel concept of hierarchical knowledge and details techniques for knowledge extraction, storage, and management across different levels. Additionally, we provide extensive experimental validation, offering a comprehensive exploration of the proposed HMI system's design, implementation, and evaluation beyond the prototype.

\section{Motivation} \label{sec:preli}
In this section, we will discuss the motivations behind our research and provide some background information. 
One goal is to provide a service that allows tenants to use customized PLMs for various tasks, such as sentiment analysis, toxic speech detection, and task-oriented dialogue systems. 
However, since many cloud service tenants do not send infer requests intensively \cite{bergsma2021generating}, 
it is not economical to provision a dedicated model instance for each task. 
Therefore, another goal is to follow the overselling strategy of the cloud -- to serve as many tenants as possible, each requiring a different PLM instance, using the least hardware requirements (e.g., a single commodity server with only one GPU).

We here analyze existing schemes and their limitations to serve a massive number of PLMs, with experiments conducted on an NVIDIA Quadro P5000 with 16GB Graphic RAM\footnote{
Service providers can choose GPUs with larger VRAM or better performance to serve more users more efficiently. However, different GPU types and VRAM sizes still face the same limitations at certain levels of workload.}.

\begin{figure}[t]
    \includegraphics[width=\linewidth]{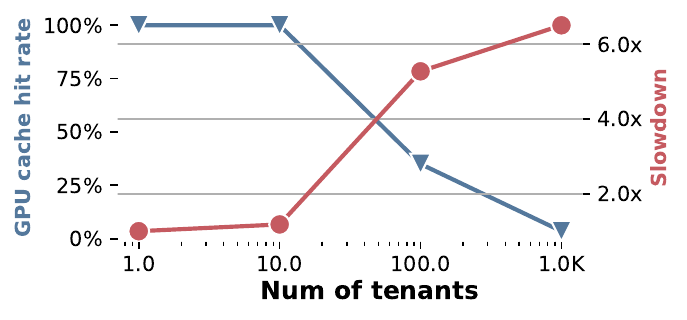}
    \caption{Impact of model swapping on GPU cache hit rate and inference efficiency slowdown ratio while serving different numbers of tenants.}
    \label{motivation:model_swapping}
\end{figure}

\paragraph{\textbf{Model Swapping.}} One straightforward solution to the problem of models that do not fit into GPU memory is model swapping. 
This method involves storing the most commonly used or recently accessed model in the GPU memory, while keeping the remaining models in host memory or on SSDs.
When an \texttt{infer} request arrives, if the relevant model is not found in GPU memory, it will be loaded from host memory to GPU for processing; 
At the same time, the least frequently accessed or least recently used model in the GPU memory is popped out to make room for the newly requested model.
However, this method may incur \textit{expensive I/O overhead}.
While this method is effective when the 
number of tenants is small, as most
requests can find their target models in GPU,
it becomes less efficient when the number of tenants increases. 
In such situations, 
as shown in \cref{motivation:model_swapping},
with 1K tenants, 97\% of requests require loading entire models before processing.
This leads to a significant reduction in inference efficiency since the wall-clock time of data transfer is typically longer than that of GPU computation. 
The \textit{slowdown} is defined as the ratio between the wall-clock time to process an inference request with model swapping and the time taken when the model resides fully in GPU memory. To evaluate GPU cache performance without bias, the workload is randomly sampled among tenants, ensuring equal access probability for each model.



\begin{figure}[!t]
    \includegraphics[width=\linewidth]{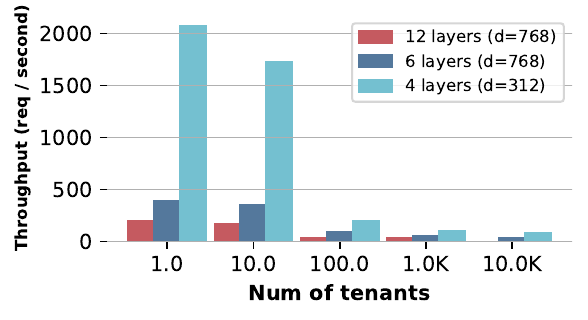}
    \caption{Impact of model compression on throughput while serving different numbers of tenants.}
    \label{motivation:model_compression}
\end{figure}
\begin{figure}[!t]
\includegraphics[width=\linewidth]{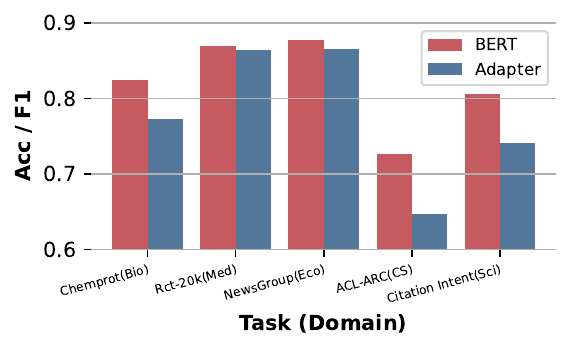}
  \caption{Performance comparison (Acc/F1) of different domain-specific tasks using parameter-efficient fine-tuning via adapter versus fully fine-tuning.}
\label{motivation:peft}
\end{figure}
\begin{figure*}[!ht]
    \centering
    \includegraphics[width=0.9\linewidth]{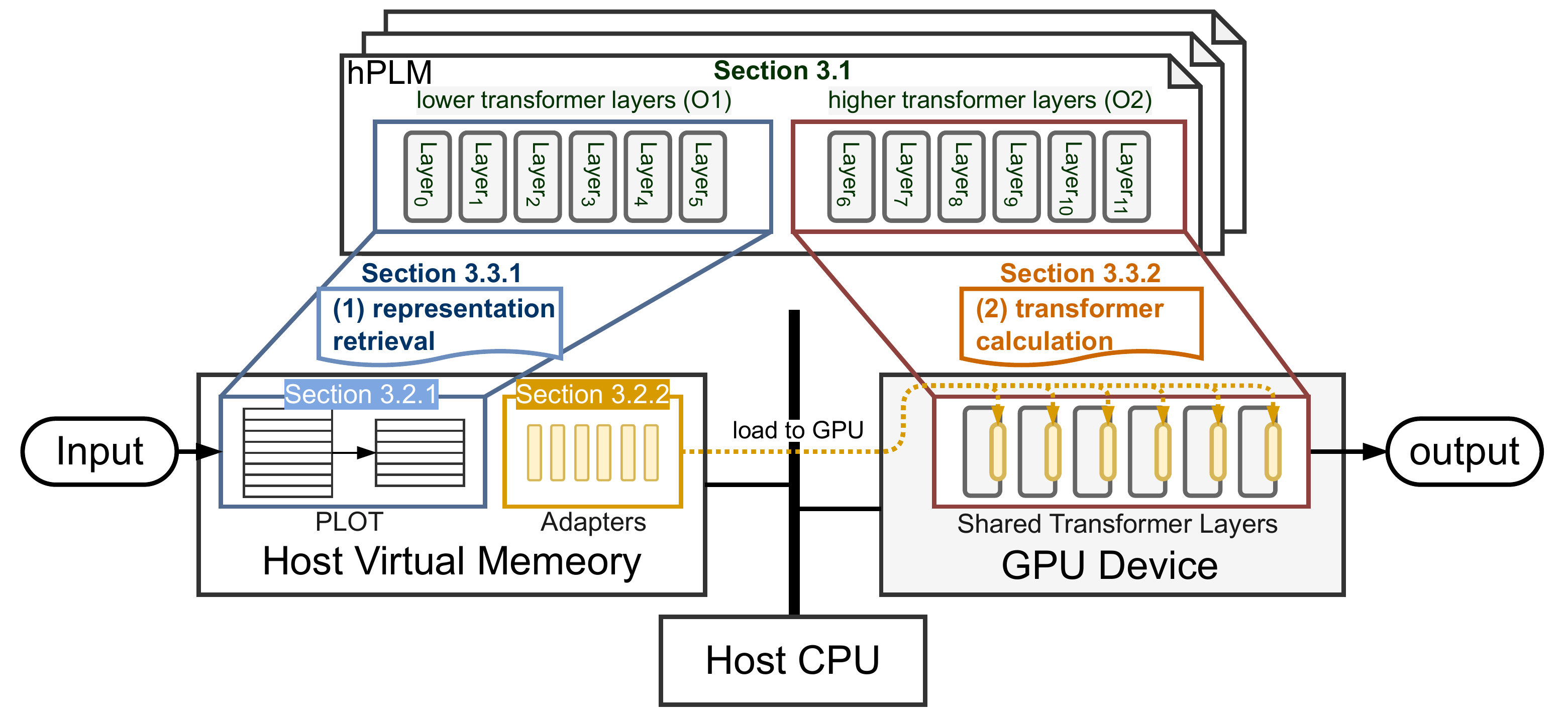}
    \caption{
    An illustration of \textsf{hPLM}s. The upper half shows the abstraction of \textsf{hPLM}s from the tenant's perspective, modeled as stacked transformer layers similar to a typical PLM. These layers are divided into lower (blue box) and higher (red box) layers, corresponding to two key observations (O1 and O2) from~\cref{sec:knowledge}, which guide hierarchical knowledge extraction. The lower half illustrates the physical data structures on hardware for hierarchical knowledge storage (\cref{sec:storage}), including PLOT (\cref{sec:plot}) and Adapter (\cref{sec:adapter}). It also visualizes (from left to right) the inference workflow (\cref{sec:infer_hplm}) for a \texttt{infer} request, comprising representation retrieval (\cref{sec:skip}) and transformer computation (\cref{sec:transformer_compute}).}



    \label{fig:hPLM}
\end{figure*}
\paragraph{\textbf{Model Compression.}}
Model compression is a technique used to reduce the size of deep learning models, thereby saving computational resources and memory.
This technique involves reducing the number of parameters in the model or compressing the model weights, 
resulting in a smaller model.
The smaller size of the model means that it can be stored and executed more efficiently, making it ideal for applications with limited computational resources.

While model compression can help to address some memory constraints, it does not sufficiently support a large number of models.
Consequently, model swapping may still be necessary to tackle memory constraints.
So in this case, this approach inherits the limitations of model swapping -- the additional communication overhead dominates, 
as shown in \cref{motivation:model_compression}.
Specifically, when the number of tenants increases to 100, solely utilizing a compressed 4-layer model (TinyBert$_4$ \cite{jiao2020tinybert}) for inference services results in a sharp decline in throughput.
Furthermore, another issue is that it can lead to \textit{a decrease in accuracy}.
By reducing 
parameter size
or compressing the model weights, some information can be lost, which can negatively impact the model's ability to make accurate predictions. 
This can be particularly problematic in applications where high accuracy is critical.
\paragraph{\textbf{Parameter-Efficient Fine-tuning.}}
Parameter-effi\-cient fine-tuning, e.g.,~adapter\cite{houlsby2019parameter}, is a technique used to fine-tune pre-trained models on new data to improve model performance while minimizing the number of parameters that need to be updated. 
This approach proves especially beneficial when fine-tuning and deploying multiple models derived from the same PLM.

One limitation is it relies on assumptions about \textit{the distribution of the new data}.
If the new data is significantly different from the pre-training data, parameter-efficient fine-tuning may result in a decrease in accuracy.
The \cref{motivation:peft}
compares the downstream performance of domain-specific tasks between fully fine-tuning and adapter.
Specifically, compared to fully fine-tuning, adapters exhibit a noticeable decline in performance (Acc/F1), particularly in tasks such as the `Chemprot' task in the `Biology' domain (represented as Chemprot(Bio)), ACL-ARC (CS), and Citation Intent (Sci) \footnote{Details of tasks across different domains are provided in \cref{ssec:setup}.}.
Moreover, while the number of parameters that need to be updated is reduced, the \textit{inference computation} is still significant. 
This can be a problem in cases where computational resources are limited or where real-time processing is required.


\section{hPLM Construction} \label{sec:vbert}


\begin{figure*}[!t]
    \centering
    \begin{minipage}{0.66\linewidth}
        \includegraphics[width=\linewidth]{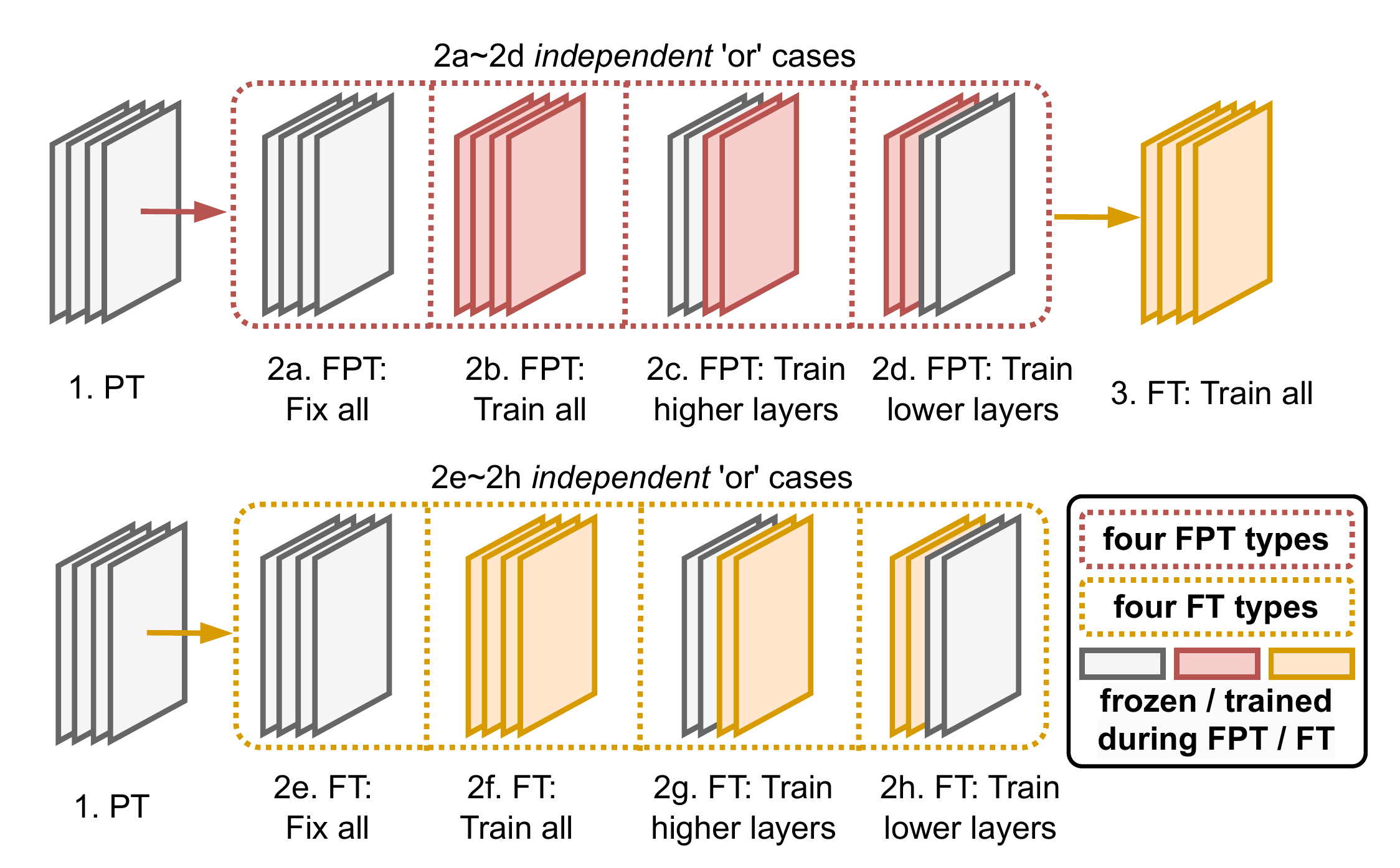}
    \end{minipage}
    \begin{minipage}{0.33\linewidth}
    \includegraphics[width=\linewidth]{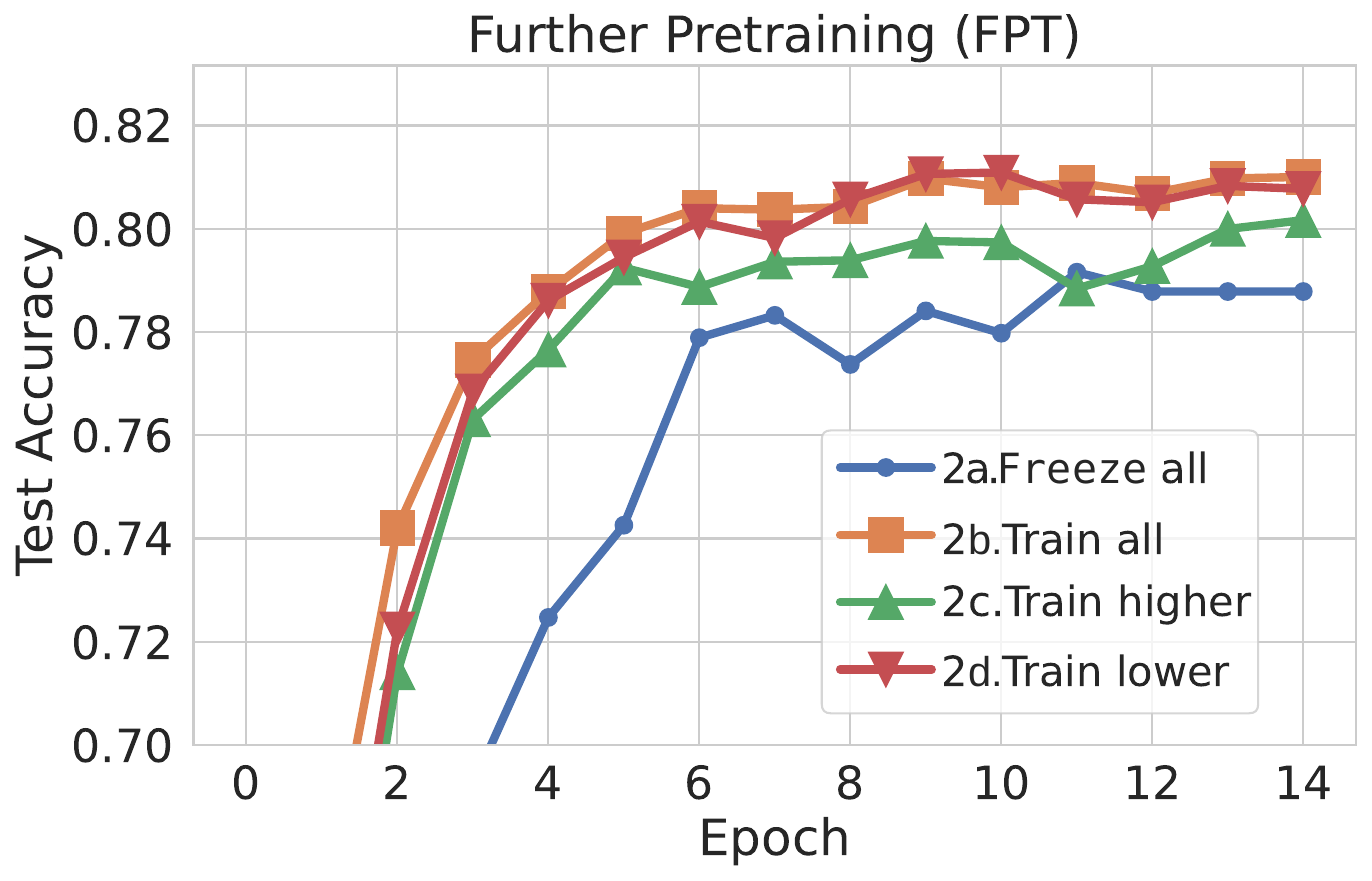}

    \includegraphics[width=\linewidth]{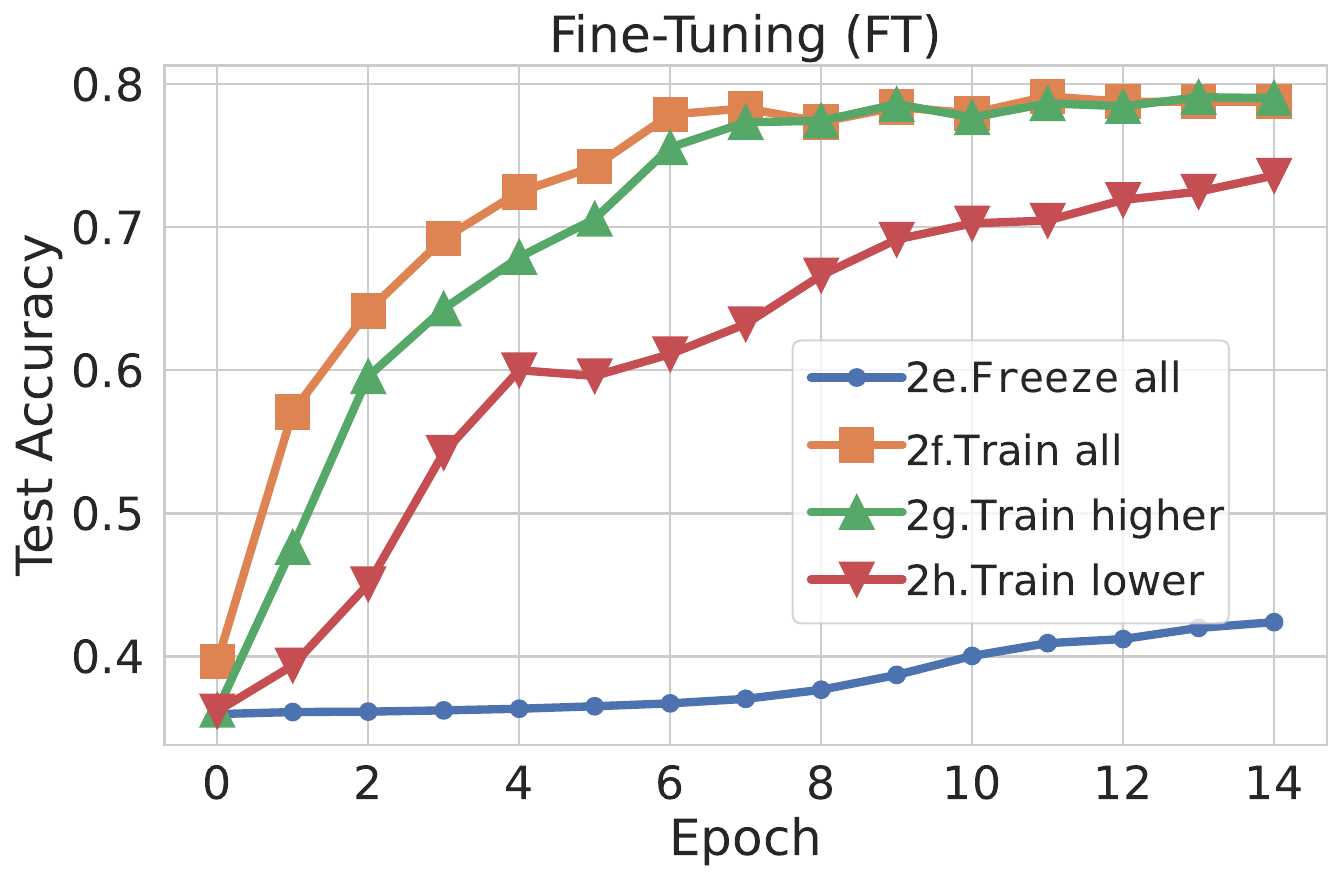}
    \end{minipage}
    
    \caption{
 \changeone{Test results by further pretraining on a domain-specific corpus before fine-tuning (upper) or by fine-tuning different subsets of layers on downstream tasks (lower).
All variants (2a to 2h,  treated as \textit{independent} `or' cases) target modifications to transformer layers while consistently training the model's output layer. As a result, even the ``Freeze all" variants (2a and 2e) exhibit performance improvements over iterations.}
 }
    \label{fig:fine-tune}
\end{figure*}
In this section, we present the \textsf{hPLM}s.
hPLMs are customized model instances for downstream tasks, constructed from pretrained models (e.g.,~encoder-only BERT or decoder-only GPT).

To create a \textsf{hPLM}, a tenant can choose either the root version, 
i.e.,~the original pretrained BERT, or branched versions, i.e.,~models that are further pretrained on domain-specific corpora, and then adapt them to the tenant's downstream datasets.
The design of \textsf{hPLM} should enable it to have similar functionality to a fine-tuned BERT, specifically, similar end-to-end accuracy, while requiring minimal hardware resource consumption.

To satisfy these requirements,
we propose to separate domain and task-specific knowledge from \textsf{hPLM}s and manage them independently to better share hardware resources. 
\cref{fig:hPLM} presents the\rone{R1.\\C8} overview of a \textsf{hPLM}.
The upper part of \cref{fig:hPLM} describes the equivalent network layers for a \textsf{hPLM} from a tenant's perspective, 
i.e.,~we can regard a \textsf{hPLM} as a stack of fine-tuned transformer layers.
The lower part of \cref{fig:hPLM} shows the physical data structures on hardware and the processing steps of tenant requests. 

In the following subsections, we first illustrate that the knowledge learned by a customized PLM is structurally separable, and therefore can be extracted hierarchically. 
Then we describe two data structures to store
this knowledge. 
Subsequently, we outline the inference steps of an \textsf{hPLM}.
Finally, we provide a step-by-step guide to construct and infer a \textsf{hPLM} from a given PLM.

\subsection{Hierachical Knowledge Extraction in PLM} \label{sec:knowledge}

Like other neural models, PLM learns knowledge from its training data by updating its neural weights. 
We categorize the knowledge learned by a fine-tuned PLM into \textit{generalized knowledge}, \textit{domain-specific knowledge}, and \textit{task-specific knowledge}. 
The generalized knowledge can be obtained when PLM is pretrained on broad and domain-agnostic corpora.
The domain-specific knowledge, which captures domain-specific language structure and syntax, is acquired in further pretraining on specialized corpora \cite{huang2019clinicalbert,lee2020biobert,scibert}, e.g.,~biological and economic corpus.
The task-specific knowledge, which describes task-specific goals and strategies, is learned when PLM is fine-tuned to downstream datasets.

    
    
An interesting finding is that the knowledge learned by \textit{further pretraining} and \textit{fine-tuning} largely pertain to different transformer layers. 
Taking BERT as an example, the domain-specific knowledge is mostly captured by the lower layers of BERT, 
and the higher layers are domain-agnostic and do not benefit much from further pretraining on domain-specific corpora.
As illustrated in the upper half of \cref{fig:fine-tune}, 
where we start from a pretrained BERT and further pretrain it on a biological corpus, 
we find training only the lower six layers of BERT (illustrated as 2d \textit{train lower} in \cref{fig:fine-tune}), i.e., updating only the neural weights in the lower transformer layers, can effectively improve the test result, whereas updating the higher six layers (illustrated as 2c \textit{train higher} in \cref{fig:fine-tune}) is marginally useful.
On the other hand, the lower half of \cref{fig:fine-tune} shows that the fine-tuning phase mostly updates the higher transformer layers (illustrated as 2g \textit{train higher} in \cref{fig:fine-tune}), whereas refining the lower layers (illustrated as 2h \textit{train lower} in \cref{fig:fine-tune}) makes a trivial contribution to the accuracy.
The above results are also confirmed on other domains/tasks, and can be generalized into two critical observations:
\begin{enumerate}
    \item[O1] 
    To achieve the utility goal of domain-specific further pretraining, it suffices to update only the lower transformer layers while freezing the higher layers during FPT. 
    \item[O2] 
    To achieve the utility goal of fine-tuning, it suffices to only update the higher layers and freeze the lower layers during FT.
\end{enumerate}
Therefore, it is possible to extract these two kinds of knowledge by splitting the transformer layers of PLM into lower and higher parts.
To facilitate efficient storage and computation of PLM, we extract the domain-specific and task-specific knowledge separately from the lower and higher layers of a PLM instance and store them in two distinct data structures to construct a \textsf{hPLM} with hierarchical knowledge.
The next subsection describes these two data structures to store hierarchical knowledge in PLM.

\begin{figure}[!t]
    \centering
\includegraphics[width=1.0\linewidth]{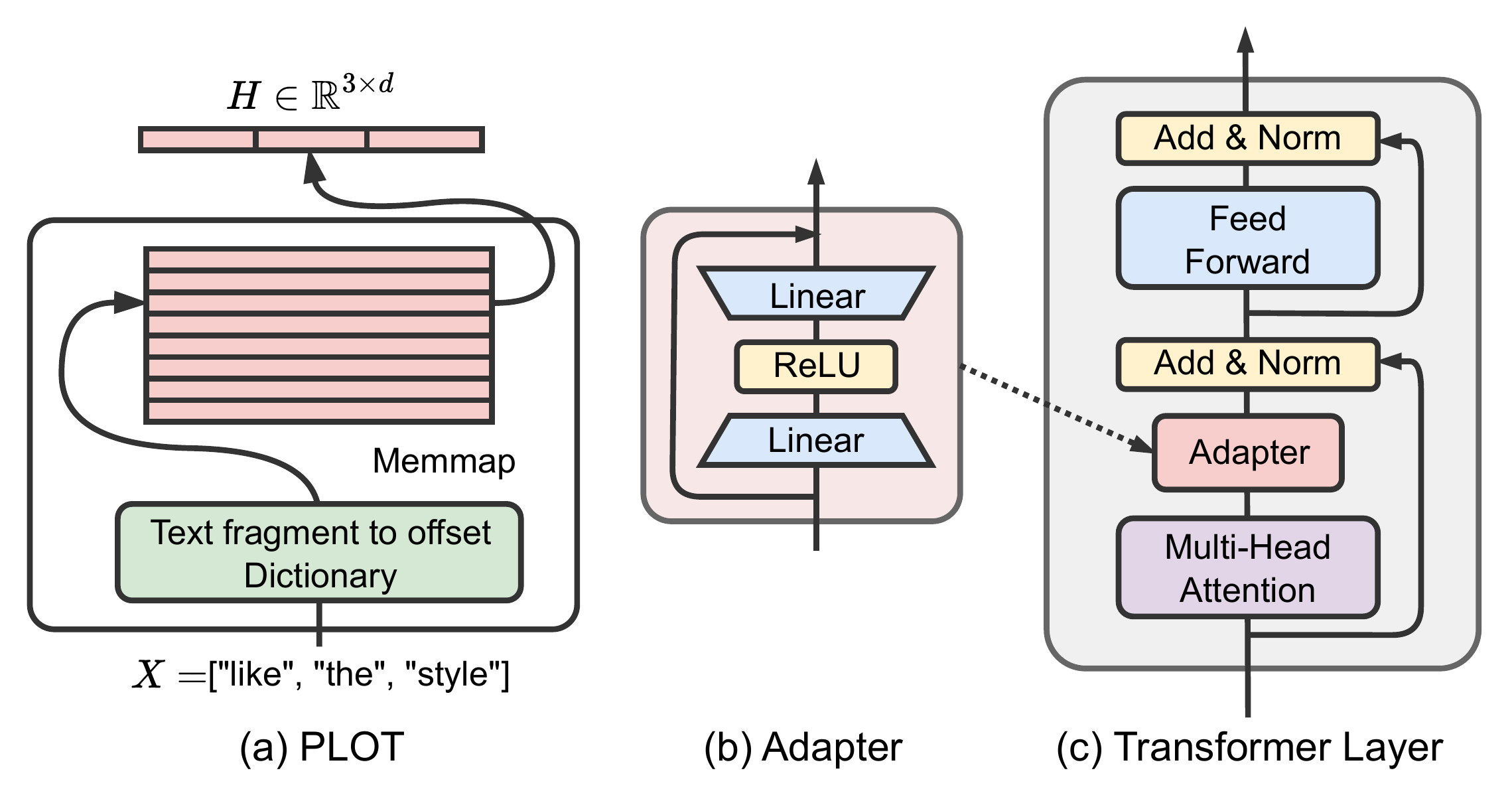}
    \caption{Data Structures of (a) PLOT, (b) Adapter, and (c) Transformer.}
    \label{fig:ds}
\end{figure}
\vspace{-5mm}
\subsection{Hierachical Knowledge Storage in PLM}\label{sec:storage}

For each \textsf{hPLM}, 
we storage 
the domain-specific and task-specific knowledge
into 
two respective data structures, namely a pretrained lookup table (PLOT) \cite{wang2022skipbert} and a set of adapters \cite{houlsby2019parameter}. 
The former is used to implement the functionality of the lower layers of PLM, while the latter can be integrated with PLM's transformer layers to comprise the higher layers of a \textsf{hPLM}. 
Both data structures are space-efficient. The PLOT is shared per domain.
The adapters are task-specific, but contain only a very limited number of parameters, which is about 3\% of the PLM parameters.

With careful design of the data structures and algorithms using them, the proposed scheme can strike a good balance between effectiveness and efficiency.

\subsubsection{PLOT}\label{sec:plot}
\changeone{\rone{R1.\\C10}\Cref{fig:ds}~(a) illustrates the data structure of PLOT,  
a key-value store where each key represents a short text fragment (an $n$-gram of tokens),  
and the value corresponds to its activation, referred to as ``representation vectors.''  
The representation of each possible short text fragment is generated by precomputing the lower transformer layers of the PLM \changeone{(i.e., the activation after these transformer layers)}.  
Notably, since the representation has already been processed through all these lower transformer layers,  
the computation of these layers is skipped during the actual inference process to improve efficiency.  
During inference, the \textsf{hPLM} searches the PLOT to retrieve text representations without the need to compute these transformer layers.  
For example, given a three-token text fragment $\v X$ = [``like'', ``the'', ``style''],  
PLOT maps it to embedding vectors $\v H \in \mathbb{R}^{3 \times d}$, where $d$ is the hidden size of the PLM.}  

\changeone{Unlike the standard computation of transformer layers, which considers all relevant tokens in the input, the computation of output activations in PLOT involves a fixed \textit{contextual} window of tokens (e.g,~tri-gram).  
For example, in the self-attention mechanism~\cite{self_attn}, encoder-only PLMs rely on tokens from both sides of a given token,  
whereas decoder-only PLMs depend on all preceding tokens.  
Thus, the embedding vectors $\v H \in \mathbb{R}^{3 \times d}$ are partially contextualized text representations, mimicking the intermediate representations of the PLM while differing from the original outputs.}
 


\subsubsection{Adapters}\label{sec:adapter}
\cref{fig:ds} (b) shows the architecture of an adapter.
Adapters are small neural structures that collect all the parametric updates during fine-tuning, while leaving the pretrained model weights, i.e.,~BERT, unchanged.
It is considered a condensed form of task-specific knowledge for a specific task, 
which can be injected into the general-purpose transformer layers to adapt them to a specific task, 
as shown in \cref{fig:ds} (c).

An adapter module is composed of two feed-forward layers with a skip connection:
\begin{align}
    \text{A}(\v a) = \text{Linear}^{\uparrow}(\phi(\text{Linear}^{\downarrow}(\v a))) + \v a
\end{align}
where $\text{Linear}(\cdot)$ is a fully-connected layer defined as 
$\text{Linear}(\v a) = \v W \v a + \v b$;
$\downarrow$ and $\uparrow$ indicate that the linear layer down-projects and up-projects the input, respectively;
$\phi(\cdot)$ is an activation function, and here we use ReLU.
\changeone{Specifically, given an input $\v a$, $\text{Linear}^{\downarrow}(\cdot)$ projects\rone{R1.\\C11,C14} it to a compact representation, while $\text{Linear}^{\uparrow}(\cdot)$ projects the activated representation using $\phi(\cdot)$ back to the original size of $\v a$, enabling the activation output to update $\v a$~\cite{houlsby2019parameter}.  
Such an adapter module with a bottleneck structure can inject task-specific knowledge into the internal activation $\v a$ of a transformer layer in a parameter-efficient manner.
A transformer layer together with an adapter module, as shown in \cref{fig:ds} (c), can be viewed as a fine-tuned transformer layer of a \textsf{hPLM} for a downstream task.  
In a \textsf{hPLM}, we follow the HuggingFace\footnote{\changeone{\url{https://github.com/adapter-hub/adapters}}} implementation to inject an adapter after the attention layer within each higher transformer layer (defaulting to the latter half) based on observation O2 in~\cref{sec:knowledge}.}

Adapters allow us to use only about 2\% additional parameters to achieve downstream accuracy similar to full fine-tuning, i.e.,~updating all parameters of BERT during fine-tuning.
Such property is desirable for deploying multiple BERT models in memory-constrained environments, 
since we only need to keep one backbone BERT and multiple sets of task-specific adapters.
By switching the parameters of the injected adapters, we can get a customized BERT model for a specific task.
\rone{R1.\\C14}\changeone{During inference, only the higher transformer layers of the \textsf{hPLM} reside on the GPU, while task-specific adapters are stored in the host's virtual memory and dynamically loaded on-demand. This design is adopted because, during fine-tuning, these layers' weights remain frozen, and only the adapter modules within these layers are updated. Consequently, for inference, the shared PLM layers can be reused across different tenants without modification.}


\subsection{\textcolor{blue}{Inference Using \textsf{hPLM}}}\label{sec:infer_hplm}

Inference of \textsf{hPLM} is conducted in two phases,
namely \textit{representation retrieval}
and \textit{transformer computation},
as shown in the lower half of \cref{fig:hPLM}.
The representation retrieval phase approximates the computation results of the lower transformer layers of BERT\footnote{
In the paper, we treat the first 6 layers of BERT-base as the lower transformer layers (and the remaining 6 layers as the higher layers).
For other configurations of BERT, we can change the number accordingly, e.g., 12 layers for BERT-large.
} in a cost-efficient way.
Instead of computing these transformer layers during inference,
our system searches for the internal representations of the input text fragments, which are precomputed and stored in PLOT.
Then the retrieved representations are aggregated to obtain the approximate internal representations from these layers.
The transformer computation phase involves a stack of pretrained transformer layers, corresponding to the upper layers of BERT, which reside in the GPU memory and are shared by all tenants.
The customization is provided by the adapter modules, fine-tuned on downstream tasks.
During inference, we transmit the respective adapters to the GPU, 
and then compute the transformer layers with these adapters installed, to adapt towards the downstream task that the \textsf{hPLM} was fine-tuned for.
We present the two phases in detail in the following.


\subsubsection{Representation Retrieval} \label{sec:skip}
We define the input text as a sequence of tokens $\v x = [x_i]_{0\le i < n}$, where $n$ is the number of tokens of the input text.
\changeone{\rone{R1.\\C13}Instead of feeding the input to the lower transformer layers of BERT, 
we sweep over the input text to get three-token fragments ($\v X_i = [x_{i-1}, x_i, x_{i+1}]$), 
and use them to search their representations from PLOT, denoted as $\v H_i, 0\le i < n$.
Larger token fragments increase search and storage costs, while smaller ones reduce context coverage, compromising performance. Following \cite{wang2022skipbert}, we default to using tri-grams to obtain reliable text representations with limited search and storage costs.}
We then approximate the outputs of the lower transformer layers of BERT  with representations of those text fragments:
\begin{align}\label{equ:agg}
    \v h &\simeq \text{Aggregate}([\v H_i]_{0\le i < n}),
\end{align}
where $\text{Aggregate}(\cdot)$ is a normalized sum of overlapping token representations in different text fragments.
\changeone{Since each token appears in three consecutive tri-grams, the aggregate function consolidates token-level information into a compact and meaningful representation by considering the token’s position within the tri-grams.}
For example, the given text ``... I really like the design ...'',  will be chunked into text fragments, including 
[``I'', ``really'', ``like''], 
[``really'', ``like'', ``the''],
and [``like'', ``the'', ``design''].
Then we sum the token representations of ``like'' in the three text fragments
and normalize it to obtain the contextualized representation of ``like''. 

The internal text representation can be approximated by short fragments, which allows us to:
1) improve the inference speed by precomputing those short fragments in advance and searching for them in inference;
2) provide customization by maintaining multiple versions of representations for different domains of tasks.

\subsubsection{Transformer Computation}\label{sec:transformer_compute}

The system maintains higher transformer layers in GPU, shared by all \textsf{hPLM}s, corresponding to the higher BERT layers, to leverage the global context of the input text.
The customization is provided by the adapter modules. 

Specifically, after obtaining the intermediate text representations from the representation retrieval phase,
a \textsf{hPLM} begins to compute the transformer layers with adapter modules installed to get the fully contextualized text representation.
For the text classification task, following \cite{bert}, we predict on the \texttt{[CLS]} token, a special token at the beginning of each sentence, and select the class label with the highest predicted score.
For the sequence tagging task, we use a linear layer to predict each token, and for each token select the tag with the highest prediction score.

\subsection{\changeone{Step-by-Step Guide to \textsf{hPLM}}}

\changeone{\rone{R1.\\C12}This section provides a step-by-step guide on constructing and performing inference with a typical \textsf{hPLM}, starting from a pretrained large language model (PLM).}

\subsubsection{\changeone{Steps of \textsf{hPLM} Construction}} 
\begin{itemize} 
    \item \changeone{\textbf{Step 1: Further-pretrain the lower layers.} 
      As outlined in~\cref{sec:knowledge} (O1), further pretrain (FPT) the lower transformer layers of the PLM (typically the first half) using domain-specific corpora while keeping the higher layers frozen. This process produces a domain-specific PLM.}

    \item \textbf{\changeone{Step 2: Fine-tune the higher layers.}} 
      \changeone{Building on~\cref{sec:knowledge} (O2), fine-tune (FT) the higher transformer layers of the domain-specific PLM (typically the second half) using task-specific data with lightweight adapters while keeping the lower layers fixed, yielding a task-specific PLM.}

    \item \textbf{\changeone{Step 3: Build domain-specific PLOT.}} 
      \changeone{As described in~\cref{sec:plot}, compute internal representations for text fragments (e.g., $n$-grams) using the domain-specific PLM applied to domain-specific corpora. These representations are used to construct the domain-specific PLOT, which is stored in host virtual memory for later retrieval in inference.}

    \item \textbf{\changeone{Step 4: Save task-specific adapters.}} 
      \changeone{Following~\cref{sec:adapter}, store the task-specific adapters obtained during fine-tuning in host virtual memory for later transformer computation in inference.}
\end{itemize}

\changeone{After these steps, the \textsf{hPLM} is fully constructed. It is important to note that hierarchical knowledge extraction and storage are performed only once for each domain and each task. Thus, these processes can be scheduled during off-peak hours. Therefore, within our system HMI (introduced in~\cref{sec:system}) primarily focuses on multi-tenant inference performance.}

\subsubsection{\changeone{Steps of \textsf{hPLM} Inference}} 
\begin{itemize} 
    \item \changeone{\textbf{Step 1: Retrieve representations from PLOT.} }
      \changeone{According to~\cref{sec:skip}, segment the input sequence into $n$-grams. Retrieve the corresponding representations from the PLOT and reconstruct the internal text representation for each token using~\cref{equ:agg}. These token representations replace the computations of the lower transformer layers of the \textsf{hPLM}, enabling efficient inference.}

    \item \changeone{\textbf{Step 2: Compute with higher layers with adapters.} }
      \changeone{As specified in~\cref{sec:transformer_compute}, load the task-specific adapters into the GPU and use them alongside the higher transformer layers of the original PLM, which also reside on the GPU. Together, these components process the token representations to generate the final output.}
\end{itemize}

\changeone{The inference process concludes with these steps. Notably, as elaborated in~\cref{sec:optimization}, the extraction of domain-specific knowledge (PLOT) and the loading of task-specific adapters can overlap with transformer layer computations during system HMI deployment. This overlapping significantly improves multi-tenant inference efficiency.}

\section{System Design of HMI} \label{sec:system}

\begin{figure*}
    \centering
    \includegraphics[width=0.8\linewidth]{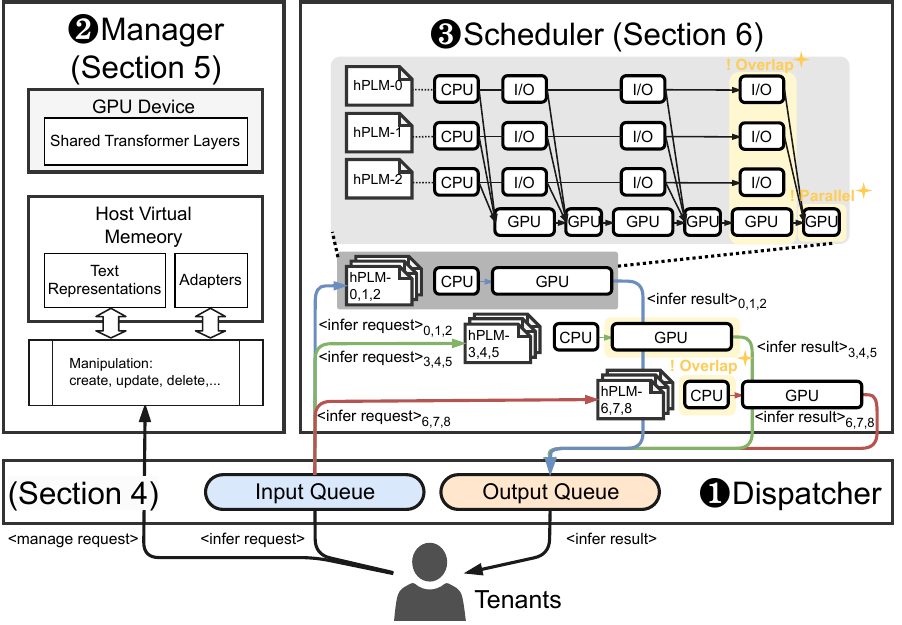}
    \caption{The System Architecture. An example with 3 inference batches, each containing 3 \texttt{infer} requests.}
    \label{fig:system}
\end{figure*}

In this section, we introduce the overall architecture of a {H}ierarchical knowledge management based {M}ulti-tenant {I}nference system, HMI, aiming to manage and deploy a large number of \textsf{hPLM} instances with limited hardware resources.
Our system is designed as a server-client architecture where the server responds to standard HTTP requests.
As depicted in \cref{fig:system}, we abstract our system into three logical components:
\ding{182} \textit{Dispatcher},
the interface between tenants and our system, 
where the labels on the arrows from tenants indicate types of requests;
\ding{183} \textit{Manager},
maintaining and manipulating \textsf{hPLM}s of tenants;
\ding{184} \textit{Scheduler},
scheduling inference jobs of \textsf{hPLM}s with fine-grained pipeline parallelism based on hierarchical knowledge prefetching to maximize hardware efficiency.

The dispatcher (as shown in \cref{fig:system} \ding{182}) receives requests from tenants, and dispatches them to the corresponding component to process them.
Next, we present the two request types handled by our system, namely the \texttt{manage} and \texttt{infer} requests.

\subsection{Manage Request}
The \texttt{manage} request ($<$manage request$>$ in \cref{fig:system}) specifies a model-related operation processed by the manager. 
We illustrate model-related operations through a lifecycle of \textsf{hPLM} as follows:
Initially, the system has one generic, pretrained PLM model, PLM-root.
To enhance its domain-specific knowledge, 
tenants could \textit{a) create a domain-specific model} by uploading a domain-specific corpus, e.g., biological data, to provide domain-specific knowledge.
On the server side, the manager uses this corpus to further pretrain PLM-root and yields a domain-specific branched version of it.
To ensure the quality and stability of the model, only domain data with a certain size of corpus can be updated to create a new branch version. 
Tenants can also \textit{b) update a domain-specific model}, i.e.,~continually pretrain existing branch models to enhance the domain-specific knowledge by uploading more data with the same domain of existing domain branches.
Next, tenants shall \textit{c) create a \textsf{hPLM} instance} by selecting an existing branch model, to acquire task-specific knowledge by fine-tuning on their customized task datasets.
Furthermore, tenants can \textit{d) update a \textsf{hPLM} instance} by continuing to fine-tune it to new data. 
Tenants can use their \textsf{hPLM} instances by sending \texttt{infer} requests,
which are handled by the scheduler to get the prediction results.
Finally, if necessary, tenants can \textit{e) delete \textsf{hPLM} instances} after use. 
\rone{R1.\\C15}\changeone{Note that if the system needs to host multiple different pre-trained models (BERT, OPT\cite{zhang2022opt}, LLaMA\cite{touvron2023llama}, or Qwen\cite{qwen}), only a simple linear extension of PLM-root is required.
Specifically, this involves constructing independent model version trees for each PLM, where each tree operates in isolation, ensuring no interference. This claim has been validated in~\cref{sec:evaluation}, where we demonstrated that independent model version trees, with roots named BERT-root and GPT-root, effectively serve \textsf{hBERT} and \textsf{hGPT} in HMI.}

\subsection{Infer Request}
The \texttt{infer} request ($<$infer request$>$ in \cref{fig:system})
specifies a \textsf{hPLM} instance to get its prediction.
The dispatcher will batch the \texttt{infer} requests (if there are multiple) and communicate with the scheduler by an input queue.
Once receiving an \texttt{infer} request, the dispatcher inserts it into the last batch of requests in the input queue.
And if the size of the last batch of requests reaches the pre-defined maximum batch size,
the dispatcher will add an empty batch to the input queue, and then insert the \texttt{infer} request to this batch.
The input queue will be synchronized to the scheduler,
which arranges the inference jobs and uses corresponding \textsf{hPLM}s to obtain the prediction results.
As presented in \cref{fig:system} \ding{183}, there are 3 batches of 9 \texttt{infer} requests with a maximum batch size of 3 arriving in the input queue in sequence, which are synchronized to the scheduler, that is, $<$infer request$>_{0,1,2}$, $<$infer request$>_{3,4,5}$ and $<$infer request$>_{6,7,8}$.
The prediction results will be placed in an output queue ($<$infer result$>_{0,1,2}$, $<$infer result$>_{3,4,5}$ and $<$infer result$>_{6,7,8}$), then the dispatcher will return those inference results ($<$infer result$>$ in \cref{fig:system}) to the corresponding tenants.

After outlining the overall framework of HMI and discussing the dispatcher's handling of two types of requests, in \cref{sec:hkm}, we delve into how the manager in HMI manages the domain and task-specific hierarchical knowledge built in multi-tenant \textsf{hPLM}s. This facilitates achieving competitive task performance within constrained storage and GPU memory resources.
Furthermore, in \cref{sec:optimization}, we elaborate on how the scheduler in HMI overlaps CPU and I/O operations with GPU operations by hierarchical knowledge prefetching and uses batched matrix multiplication to optimize parallel implementation. These optimizations enable HMI to maximize hardware efficiency for efficient batch inference.

\section{Hierarchical Knowledge Management in HMI} \label{sec:hkm}
Here, in \cref{sec:dkm}, we introduce the process of further pre-training domain-specific models, and elaborate on how the manager (as illustrated in \cref{fig:system} \ding{183}) constructs PLOT version trees by the frequency-based updating mechanism, thereby managing domain-specific knowledge with an acceptable increase in additional storage. Additionally, in \cref{sec:tkm}, we describe the process of fine-tuning task-specific models and detail how the manager employs the adapter's swapping strategy to manage task-specific knowledge under limited video memory.

\subsection{Domain-Specific Knowledge Management} \label{sec:dkm}

As presented in O1 of \cref{sec:knowledge},
we observe that FPT only needs to update the lower layers while keeping the higher layers unchanged.
Moreover, we also find that the lower layers can be materialized in a scheme called lower layer skipping, 
where the domain-specific knowledge obtained by FPT can be extracted as the text representations are stored in PLOT.
Therefore, we propose to manage domain-specific knowledge learned from FPT in the form of different versions of PLOT tables.
We here show how to perform FPT and how to manage multiple domain-specific text representations.
\paragraph{\textbf{Further pretraining domain-specific models.}}
The manager can create a domain-specific branched model by further pretraining BERT-root on domain-specific data.
We employ self-supervised learning with a masked language model (MLM) training objective and use AdamW optimizer to train the model.
Following \cite{bert}, we randomly select 15\% tokens in the input text, 
of which 80\% become \texttt{[MASK]}, 10\% are replaced by random tokens, and 10\% remain unchanged.
And we let the model predict the selected tokens, and minimize the cross-entropy loss between the prediction and ground truth,
i.e.,~the original input text without masking.
We only update the lower 6 layers while keeping the higher 6 layers unchanged.
After FPT, we use the trained model to precompute the representations of text fragments and store them in a PLOT table.

\paragraph{\textbf{Managing representation entries.}}
\changeone{\rone{R1.\\C16}For most text fragments, their respective precomputed representations do not differ significantly under different domains, 
so they can be shared across tenants which is further revealed in \cref{exp:plot} that sharing 70\% of the text representation among domains yields comparable effectiveness to using the text representations from each domain individually.}
For text fragments whose distribution and semantics differ significantly between domains, such as those containing professional terms, we keep them in a separate PLOT table.

Specifically, we maintain a PLOT table for the root version, $\Phi_{\text{root}}$,
and one for each domain-specific branch version, $\Phi_{v}$, where $v$ is the branch version.
In such a table, 
an entry refers to a key-value pair, where
the key is a text fragment, and the value is its representation.
When further pretraining is complete, the manager will create a new branch version $v'$ and use the further-pretrained model to compute the internal representations of text fragments in the domain-specific corpus, and insert them in $\Phi_{v'}$.

\changeone{\rone{R1.\\C16}Motivated by the fact that high-frequency text fragments are more likely to be domain-specific and thus more deserving of updates, whereas low-frequency fragments tend to be shared across domains and can remain in the root version. Thus, to reduce the redundant overlapping entries between the new version $v'$ and the root version,
we adopt a simple frequency-based strategy:}
The manager will only add a text fragment and its representation to $\Phi_{v'}$,
if this text fragment appears in the $\alpha\%$ of the most frequent text fragments, 
where $\alpha\%$ is a pre-defined threshold\footnote{The impact of different $\alpha\%$ on task performance and additional storage requirements be discussed in \cref{exp:plot}.}.


During inference, the system will first look up the domain-specific PLOT table, and in case of missing, look up the root version.
Formally, given a text fragment $\v X_i$ and its corresponding representation version $v$, we can obtain the representation as follows:
\begin{align}
    \v H_i =
    \begin{dcases*}
        \Phi_{v}[\v X_i],  &\text{if $\v X_i \in \Phi_{v}$}; \\
        \Phi_{\text{root}}[\v X_i], &\text{else.} \\
    \end{dcases*}
\end{align}

\changeone{\rone{R1.\\C16}For out-of-vocabulary (OOV), where a text fragment is not even found in the root version $\Phi_{\text{root}}$, we return bi/uni-gram representations as a backup strategy to pre-compute, similar to the tri-gram. Specifically, in PLOT, we append bi/uni-gram text fragments and their corresponding representations using the method outlined in~\cref{sec:plot}, as they can be easily managed due to their far smaller token combinations compared to tri-grams, e.g., uni-gram is the size of the token vocabulary.}
Notably, representation retrieval during inference occurs on the CPU. Prefetching this domain-specific knowledge can hide this operation within GPU-intensive transformer computations, thereby enhancing batch inference efficiency. Further discussion on this optimization is provided in \cref{sec:optimization}.

\subsection{Task-Specific Knowledge Management} \label{sec:tkm}

Tenants can select a branch version of BERT, and upload their specifications and data to fine-tune a customized \textsf{hPLM} instance.
Here we present how the system creates \textsf{hPLM}s and manages their task-specific knowledge. 

\paragraph{\textbf{Fine-tuning task-specific models.}}
The manager begins from a domain-specific branch model, as specified by the tenant,
and randomly initializes a set of adapters and injects them into the transformer layers, i.e.,~the shared transformer layers in this work.
During fine-tuning, we keep the transformer layers unchanged and only update the parameters of the adapters and the output layer.
We only insert adapters into the higher 6 transformer layers, since the lower 6 layers are materialized in the form of PLOT and we noted in O2 of \cref{sec:knowledge} that fine-tuning lower layers has a trivial contribution to the downstream accuracy.

\paragraph{\textbf{Managing parameters of adapters.}}
Adapters can facilitate effective model reuse, allowing us to maintain only a few megabytes of space for each downstream task from the tenant.
However, we cannot host all adapters in a single GPU when there are hundreds or even thousands of \textsf{hPLM} instances.
To address this issue, we store the parameters of the adapters in the host memory via a swapping strategy once a PLM is fully fine-tuned.
These parameters are reloaded to GPU device only when the corresponding \textsf{hPLM} is requested by the tenant.
We show in \cref{sec:optimization} that the loading time can be hidden in GPU computation and thus will not incur additional run time during inference.

\section{Implementation and Optimization of HMI} \label{sec:optimization}


Although HMI is designed to support efficient inference, we show here that the inference speed can be further improved with better scheduling schemes.

\begin{figure*}
    \centering
    \includegraphics[width=0.8\linewidth]{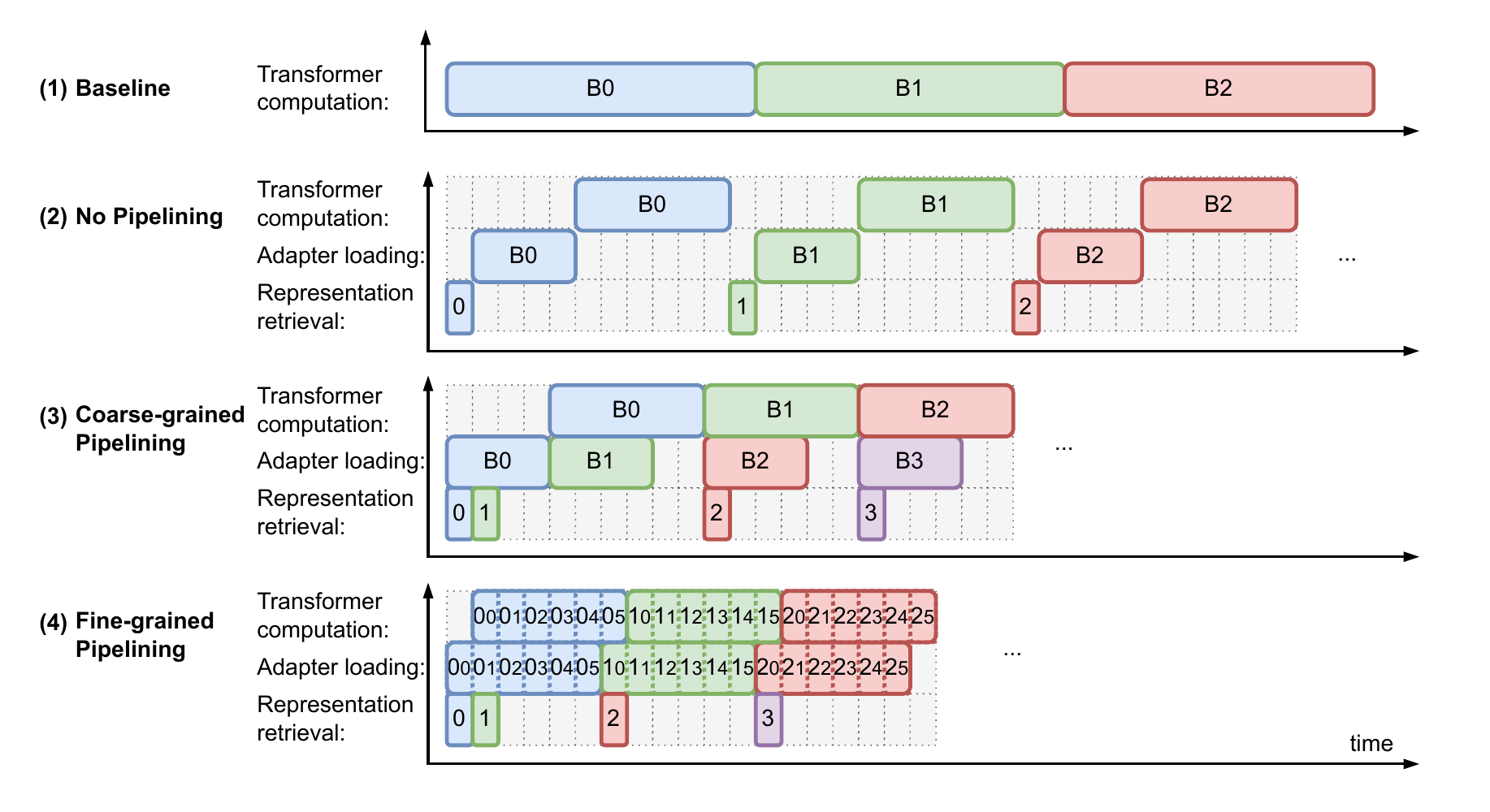}
    \caption{Comparison of inference pipeline schemes.
            B$i$ means the $i$-th batch of requests.
            $i_j$ indicates the block processing $j$-th transformer layer for batch $i$. The representation extraction of B$2$ does not follow that of B$1$. This illustrates a scenario where we wait for all requests in the batch to arrive (reaching the maximum batch size) before extracting their representations simultaneously.
    }
    \label{fig:pipeline}
\end{figure*}

First, for PLMs inference baseline with 3 batch (B) requests, it only performs GPU-intensive \textit{transformer computation} in batch arrival order, as shown in \cref{fig:pipeline} (1). However, for \textsf{hPLM}s, as shown in \cref{fig:system} \ding{184}, the forward pass of a \textsf{hPLM} instance contains a mixture of CPU, I/O, and GPU operations. Specifically, the \textit{representation retrieval} is executed by CPU,
\textit{adapter-loading} is I/O intensive,
and the \textit{transformer computation} is handled by GPU. Since it realizes lower layers computations skipping through effective extraction of domain knowledge, its inference speed is improved as shown in \cref{fig:pipeline} (2). However,
conducting inference of different requests sequentially will result in poor resource utilization 
-- the system either performs CPU, I/O, or GPU operations, while other parts are idle, leading to low hardware efficiency.
Here, the scheduler (as shown in \cref{fig:system} \ding{184}) is introduced to orchestrate the inference pipeline to improve the inference efficiency:
it aims to overlap the CPU and I/O with GPU operations by fine-grained pipeline parallelism based on hierarchical knowledge prefetching.

Moreover, due to the nature of cloud services, 
requests typically involve different model instances.
So for each model, the inference batch size is usually small, 
making the compute kernel extremely fragmental and leading to low hardware utilization.
Our HMI design introduces a shared stack of transformers, so we can batch different requests to enable higher hardware efficiency.
For the customization part, i.e.,~adapters, we present a parallel inference implementation to fuse the execution of different requests.

The implementation details will be presented in the subsequent subsections.
\subsection{Pipelined Representation Retrieval} \label{ssec:prr}

By introducing the representation retrieval phase,
we convert part of the GPU computation to workloads on CPU searching for text representations.
We note that the GPU computation time is usually much longer than searching representations
-- a single forward pass on 6 layers of BERT requires 6 ms, whereas the latter needs 0.2 ms from memory and 2.4 ms from SSD.
One can hide the overhead of searching representations within the forward pass of the previous batch of requests.

\cref{fig:pipeline} shows how to achieve pipeline parallelism by prefetching domain-specific knowledge (text representation in PLOT) to improve inference efficiency.
Specifically, for the first batch of requests (B0), 
the system shall (i) search representations from PLOT, (ii) load the parameters of adapters, and (iii) perform calculations on the GPU.  
During step (iii) of B0, CPU actually becomes idle, and is available for processing of the next batch (B1), as illustrated in \cref{fig:pipeline} (3).
In particular, since the time of representation searching is usually much smaller than GPU computation,
the former can be mostly hidden in the latter, thus improving the inference throughput.
\vspace{-5mm}
\subsection{Pipelined Adapter Swapping} \label{ssec:pas}

Although adapters containing task-specific knowledge are lightweight in computation compared to BERT,
they can still saturate the GPU memory when the number of model instances increases.
Keeping them to the host memory could resolve this problem, 
but incurs additional communication overhead in loading adapters to the GPU.
Fortunately, we note this issue can be addressed by system optimization, 
and eventually does not significantly impact the end-to-end runtime.

We first observe that the transformer layer is usually computationally heavy, 
while the adapter module is often lightweight in terms of the number of parameters, 
so the GPU computation time of the transformer is usually much longer than the data transfer time to load the adapter module.
Therefore, we can easily hide the data transfer in the GPU computation of the previous batch.
Following the similar idea of pipelining the representation retrieval,
we can load the parameters of adapters with coarse-grained pipelining, prefetching the task-specific knowledge (adapters) layer-wise for the next batch, as shown in \cref{fig:pipeline} (3).
Loading adapter parameters and GPU computations are assigned to different CUDA streams so that they will be executed in parallel; as a result, communication will overlap with computation to reduce inference time.

Moreover, to further reduce the response time, 
we can perform GPU computation of the current transformer layer and prefetch the adapter module of the next layer at the same time,
as shown in \cref{fig:pipeline} (4) the fine-grained pipelining.
Therefore, this method allows the system to utilize the host memory, which is typically much larger than GPU memory, 
to accommodate more adapter modules without slowing down the inference speed.
We by default enable fine-grained pipelining as it performs well in most scenarios.

\vspace{-5mm}
\subsection{Parallel Inference of Adapters} \label{ssec:pia}

Despite the parameter efficiency of adapter tuning,
which only incurs 2\% memory footprint of the original BERT model,
the naive implementation does not consider concurrent inference of multiple independent adapters.
Thus, conducting inference over adapters iteratively for each tenant will cause low hardware efficiency.
In this paper, we adopt the parallel implementation of adapters \cite{ruckle2020adapterdrop}.
The original goal of parallel adapters was to compute multiple off-the-shelf adapters and fuse their outputs for the same input text.  
However, we aim to compute independent inputs with different adapters in parallel.
We stack the parameters of adapters and perform batched matrix multiplication to achieve this.\footnote{
\url{https://pytorch.org/docs/stable/generated/torch.baddbmm.html}
}
Our parallel implementation proved to be consistently more efficient than the naive iterative approach as shown in \cref{exp:pipeline}.

\section{Evaluation} \label{sec:evaluation}

In this section, we empirically evaluate our approach under various scenarios. A preview of our main experiments and findings is outlined below.
\begin{enumerate}
    \item We compare the GPU memory footprint and show that our approach can host thousands of model instances ($>$10K \textsf{hBERT}s) with one GPU (see \cref{exp:memory});
    \item \textcolor{blue}{We compare the inference throughput and end-to-end response time and show that our system is efficient and can support up to 10K tenants, with approximately 1,000$\times$ end-to-end response time speedup compared to a standard PLM via model swapping (see \cref{exp:response})};
    \item On a wide range of tasks, our approach achieves the minimal accuracy drop, $\sim$1\% average accuracy reduction compared with fine-tuning all parameters of BERT (see \cref{exp:acc}). 
    \item We extend our scheme to generative PLMs ( \textsf{hGPT}s) and achieve significant throughput improvements over baselines (see \cref{sec:GPT}).
    \item We analyze the effectiveness of space and time optimization of HMI systems. For space efficiency, the PLOT version tree built with frequency updates requires only 18.9GB of additional storage for 10 domains (see \cref{exp:plot}).
    For time efficiency, fine-grained pipeline parallelism using hierarchical knowledge prefetching and adapter parameter parallelism via batch matrix multiplication improves system efficiency by 25\% (see \cref{exp:pipeline}).
\end{enumerate}

\subsection{Setup}\label{ssec:setup}

\paragraph{Environment.} All the experiments in the paper are carried out on
a server with RAM of 312GB, and an SSD of 2TB GB,
NVIDIA Quadro P5000 with 16GB Graphic RAM, all of which are connected via PCIE-3.0.
The server is powered by an Intel(R) Xeon(R) @ 2.40GHz processor with 64 cores. 
We use PyTorch 1.12 and CuPy compiled with CUDA 11.3.

\begin{table}[!t]
    \caption{Dataset Statistics. ``\textsc{CLS}'' means classification task; ``\textsc{NER}'' means named entity recognition.}
    \centering
    \small
    \setlength{\tabcolsep}{1pt}
    \begin{tabular}{@{}lllrrr@{}}
    \toprule
        FT Data      & Domain  & Task   & \# train & \# dev & \# test \\
    \midrule
        ChemProt        & Biology   & \textsc{CLS}   & 4169 & 2427 & 3469 \\
        BC5CDR          & Biology   & \textsc{NER}   & 4942 & 4949 & 5139 \\
        JNLPBA          & Biology   & \textsc{NER}   & 18607 & 1939 & 4260 \\
        NCBI-disease    & Biology   & \textsc{NER}   & 5424 & 923 & 940 \\
        RCT-20k         & Medicine  & \textsc{CLS}   & 180040 & 30212 & 30135 \\
        NewsGroup       & Economics & \textsc{CLS}   & 9054 & 3631 & 5249 \\
        ACL-ARC         & Computer Science & \textsc{CLS}   & 1688 & 114 & 139 \\
        SciERC          & Computer Science & \textsc{NER}   & 1861 & 275 & 551 \\
        Citation Intent & Bio+Med  & \textsc{CLS}   & 1688 & 114 & 139 \\
        SCI-Cite        & Bio+Med  & \textsc{CLS}   & 7320 & 916 & 1861 \\
    \bottomrule
    \end{tabular}
    \label{tab:data}
\end{table}

\paragraph{Data.}  
We evaluate a collection of classification (\textsc{CLS}) and named entity recognition (\textsc{NER}) tasks, including ChemProt, BC5CDR, JNLPBA, NCBI-disease, RCT-20k, NewsGroup, ACL-ARC, SciERC, Citation Intent, and SCI-Cite. The dataset statistics are summarized in \cref{tab:data}. Note that ACL-ARC and Citation Intent share the same input but differ in labels and task types.
\changeone{We spawn a \textsf{hBERT} instance for each task and adapt it to the corresponding task for~\cref{exp:acc}. 
Further, to construct 10K \textit{tenant-specific} \textsf{hBERT}s, where each \textit{tenant} corresponds to a distinct \textsf{hBERT} dedicated to a specific task subset, we allocate the number of tenants $n_t$ for each task $t$ in proportion to its training dataset size $|D_t|$, ensuring fairness and unbiased scaling as follows:}

\begin{align}
\changeone{n_t = \left\lfloor \frac{|D_t|}{\sum_{t \in T} |D_t|} \cdot N \right\rfloor, \quad \text{where } N = 10,000.}
\end{align}
\changeone{Each subset \(D_s^t\) ($D_s^t \subseteq D_t$) for a tenant under task $t$ is constructed through random sampling with replacement, preserving the statistical properties of the original datasets without bias, with an equal number of samples across tenants for the same task:}
\begin{align}
\changeone{|D_s^t| = \frac{|D_t|}{n_t}, \quad P(x \in D_s^t) = \frac{1}{|D_t|}, \, \forall x \in D_t.}
\end{align}
We report accuracy for classification tasks and entity-level F1 scores for \textsc{NER} tasks.

To build the root version of BERT, we use the  checkpoint\footnote{
\url{https://github.com/LorrinWWW/SkipBERT}
} released by \cite{wang2022skipbert}, containing 6 lower transformer layers (to be materialized and stored in PLOT) and 6 higher transformer layers.
We perform further pre-training on domain corpora, including Biology, Physics, Medicine, Economics, and Computers Science data, 
to obtain the branch versions (i.e.,~internal nodes in \cref{fig:branch}) and domain-specific PLOTs.
Those corpora are powered by {S}2{ORC} \cite{lo-wang-2020-s2orc}, a general-purpose data collection over scientific papers.

\paragraph{Hyperparameter Tuning.}
We conduct careful hyperparameter tuning for all methods on all the datasets. 
We perform a grid search to choose the learning rate from [5e-6, 5e-4] and batch size from [2, 128] for the best model performance on the held-out development data set.
For each run, we train 15 epochs with the AdamW optimizer, keep the model instance that achieves the highest score on the development set, and then evaluate and report the test score.

\subsection{Compared Schemes}

We mainly compare the following schemes:

\textbf{1. \textsf{BERT-dedicated}.} 
We fine-tune all parameters of BERT for each task separately.
We use the base-uncased checkpoint\footnote{\url{https://huggingface.co/bert-base-uncased}}.
During inference, we use an individual process to maintain a model instance, and the dispatcher sends \texttt{infer} requests to the corresponding process.
When the GPU cannot host all models when the number of models is large,
we keep models in host memory and swap them into GPU only when relevant \texttt{infer} requests arrive.
This could be considered as a strong baseline in terms of accuracy.

\textbf{2. \textsf{BERT-shared}.} 
A naive solution to avoid creating an independent model instance for each tenant is to use BERT as a feature extractor and only update the output layer during FT.
However, this approach might not obtain acceptable accuracy on most customized tasks.

\textbf{3. \textsf{BERT-compress}.}
We include schemes that employ model compression.
Specifically, we examine both \textit{\textsf{DistilBERT}} and \textit{\textsf{TinyBERT}}.
\textsf{DistilBERT} \cite{sanh2019distilbert} compresses the model by performing prediction-based distillation with BERT.
\textsf{TinyBERT} \cite{jiao2020tinybert} compresses the model by performing layer-wise distillation with BERT.
The 6 layer and 4 layer checkpoints are denoted as \textsf{TinyBERT}$_6$ and \textsf{TinyBERT}$_4$.
Like \textsf{BERT-dedicated}, we use an individual process to maintain each model instance. 
We by default fine-tune and compare with \textsf{TinyBERT}$_6$ as it achieves better accuracy in downstream tasks.
We adopt the model swapping strategy in case the GPU cannot accommodate all models.

\textbf{4. \textsf{hBERT}.} 
\changeone{\rone{R1.\\C19}\textsf{hBERT} is our proposed scheme,
which freezes the parameters of BERT, injects task-specific adapter modules (adapter-tuning), and materializes the lower transformer layers (PLOT-based lower layer skipping).
We also compare to \textsf{hBERT} variants with adapter-tuning only and PLOT-based lower layer skipping only, 
which are denoted by \textit{\textsf{hBERT-adapter}} and \textit{\textsf{hBERT-PLOT}} respectively.}

\subsection{Comparison of GPU Memory Usage} \label{exp:memory}

\cref{tab:memory} provides a comparison of the GPU memory usage for different schemes. 
Specifically, we report the incremental memory usage when a new model instance is added, and the maximum number of models that can be loaded into a GPU with 16GB VRAM. 
For simplicity, we exclude the parameters of the output layer, which are typically small compared to BERT (around 30KB versus 440MB).

Among the schemes presented, \textsf{BERT-shared} has good scalability since it does not require additional task-specific parameters and uses the original BERT model. 
However, \textsf{BERT-shared} may not perform well on downstream tasks, highlighting the need for customization to achieve acceptable accuracy (see \cref{exp:acc}).

On the other hand, \textsf{BERT-compress} (\textsf{DistilBERT} and \textsf{TinyBERT}) reduces the model size through knowledge distillation, 
leading to a lower per-instance GPU memory footprint. 
However, \textsf{BERT-compress} still requires an entire model copy for each downstream task, 
limiting the number of model instances that can be accommodated in one GPU simultaneously.

Our \textsf{hBERT} variants offer effective resource sharing.
\textsf{hBERT-PLOT} reduces BERT's GPU memory requirements by half, while performing on par with, or sometimes better than \textsf{BERT-dedicated} (see \cref{exp:acc}).
Meanwhile, \textsf{hBERT-adapter} reduces the per-instance GPU memory footprint by utilizing adapter-tuning. 
This approach can accommodate significantly more model instances by sharing a backbone model and a large number of adapters.
Importantly, \textsf{hBERT}'s efficient implementation of parameter swapping (see \cref{sec:optimization} and \cref{exp:pipeline}) ensures that the overhead of loading adapter parameters from host to GPU is minimized.
This allows \textsf{hBERT} to place adapter parameters at host memory, freeing up GPU memory and making it possible to support a larger number of model instances. 
When enabling both techniques, \textsf{hBERT} can easily support over 10K model instances with only a single GPU,
with only minor accuracy degradation.

\changeone{\rone{R1.\\C21}Notably, HMI does not involve inter-GPU optimizations and is inherently GPU-agnostic. As long as GPU memory suffices to store the upper layers of individual base models as backbones, HMI achieves linear scalability. In contrast, compressed models face significant challenges even in multi-GPU setups. For instance, as shown in~\cref{tab:memory}, a 16GB GPU can support up to 60 tenants, whereas matching HMI's capacity would require approximately 167 GPUs—a prohibitive cost for cloud service providers. Furthermore, \cref{exp:acc} shows that \textsf{BERT-compress} underperforms \textsf{hBERT} in downstream tasks, further limiting its viability.}

\begin{table}[t]
\caption{
    Comparison of GPU memory footprint.
    *: they remove per-instance parameters from GPU memory to host memory, so the maximum number of instances is not limited by the size of GPU memory.
    }
\small
\setlength{\tabcolsep}{0.8pt}
\centering
\begin{tabular}{@{}lrrrrrrrrc@{}}
\toprule
Method      
            & \begin{tabular}{@{}c@{}} GPU Mem \\ base \end{tabular}
            & \begin{tabular}{@{}c@{}} GPU Mem \\ per instance\end{tabular}
            & \begin{tabular}{@{}c@{}} Max \# instances \\ (16GB VRAM) \end{tabular}
            \\
\midrule   
\textsf{BERT-dedicated}        
            & 418 MB    & 418 MB  & 35     \\
\textsf{BERT-shared}   
            & 418 MB    & -      & {$>$10,000} \\
\textsf{BERT-compress}\textsubscript{distil}
            & 253 MB    & 253 MB  & 60    \\
\textsf{BERT-compress}\textsubscript{tiny6}
            & 253 MB    & 253 MB  & 60      \\
\textsf{BERT-compress}\textsubscript{tiny4}
            & {55 MB}  & 55 MB   & 280    \\
\hline
\textsf{hBERT-PLOT}    
            & {162 MB}    & 162 MB  & 90   \\
\textsf{hBERT-adapter}*     
            & 418 MB    & {(6 MB)}  & {$>$10,000}  \\
\textsf{hBERT}*      & {162 MB}    & {(3 MB)}  & {$>$10,000}  \\
\bottomrule
\end{tabular}
\label{tab:memory}
\end{table}

\subsection{Comparison of Inference Efficiency} \label{exp:response}

In this subsection, we measure and compare \textit{throughput} and \textit{end-to-end response time} with different number of tenants,
as well as \textit{response time of burst requests}.
In this section on efficiency experiments related to \changeone{video RAM (VRAM) usage,}\rone{\ \ \ \ \ \ \ R1.\\C20} we select the variants with lower GPU memory consumption for comparable methods. According to \cref{tab:memory}, for \textsf{BERT-compress}, we use \textsf{BERT-compress}\textsubscript{tiny4}; For our method, we select \textsf{hBERT-adapter} and \textsf{hBERT}.

\begin{figure}[!t]
\centering
\includegraphics[width=1.0\linewidth]{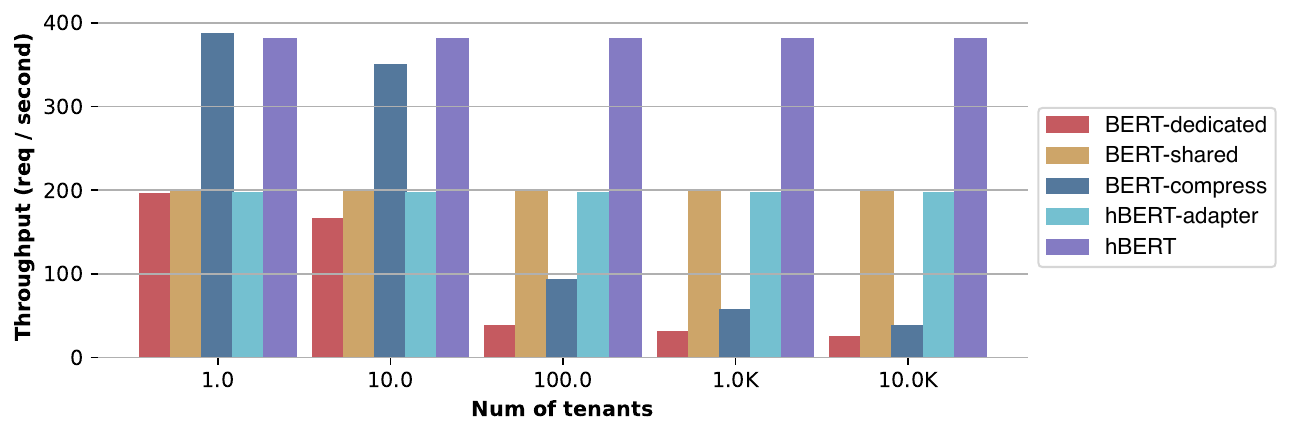}
\caption{Throughput with different number of tenants. }
\label{fig:throughput}
\end{figure}


\paragraph{Throughput.}
We calculate the throughput of an inference batch by
$\mathbb{E}[\frac{\text{batch size}}{\text{infer time of a batch}}]$, 
where an inference batch is set to be reasonably large to maximize the inference efficiency.
And the sequence length is set to 128. 
However, it is important to note that, in the cloud setup,
\texttt{infer} requests often come from various tenants and their target models are different.
In such cases, for \textsf{BERT-dedicated} and \textsf{BERT-compress} models, we split the batch into smaller mini-batches and send them to the corresponding model for processing.

Our evaluation of inference throughput is presented in Figure \ref{fig:throughput}. 
For \textsf{BERT-dedicated} and \textsf{BERT-compress}, we use the model swapping strategy when the number of models exceeds the space capacity of the GPU. 
Unfortunately and yet expectedly, as shown in the figure, the throughput significantly decreases along with the number of tenants, and inference thus becomes communication bound.

In contrast, \textsf{BERT-shared}, \textsf{hBERT-adapter}, and \textsf{hBERT} all share a backbone model across tenants. 
As a result, these models can efficiently batch the \texttt{infer} requests, maintaining a constant throughput even as the number of models or \texttt{infer} requests increases.
However, it's important to highlight that even with fewer tenants (1 or 10), \textsf{hBERT} sustains either optimal or suboptimal throughput.
\begin{figure}[t]
\centering
\includegraphics[width=1.0\linewidth]{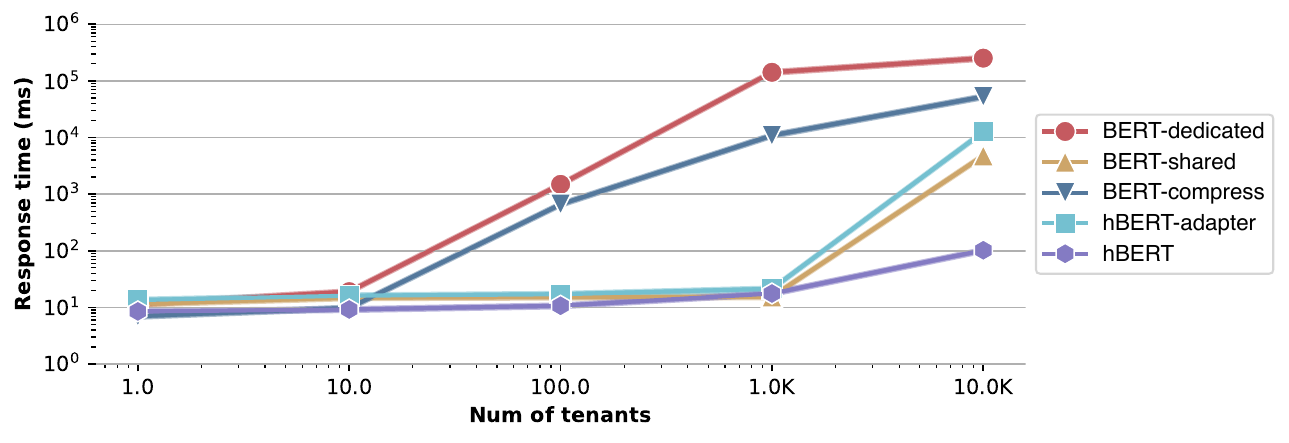}
\caption{End-to-end response time with different number of tenants (results are plotted in logarithm-scale).}
\label{fig:response_time}
\end{figure}

\begin{figure}[t]
    \centering
    \includegraphics[width=1.0\linewidth]{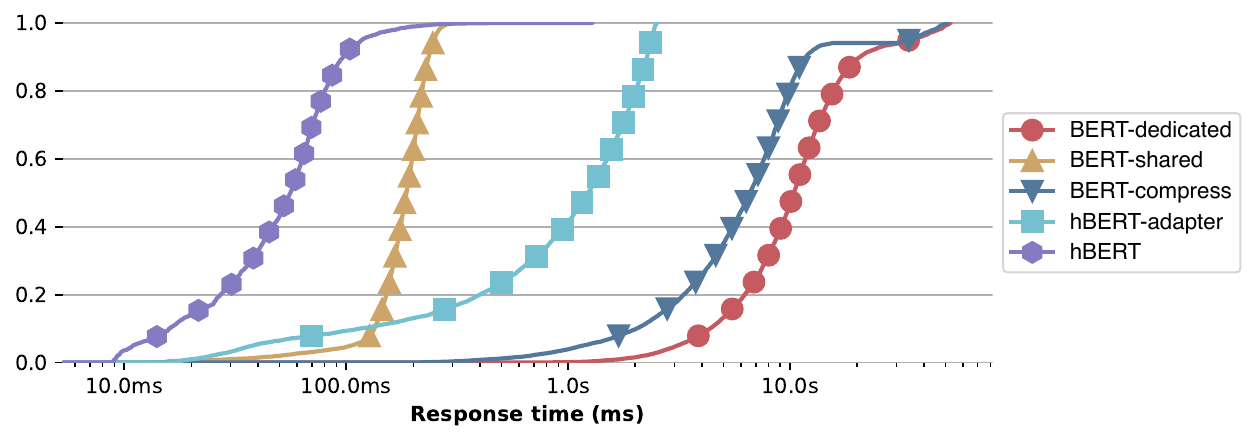}
    \caption{Cumulative distribution of response time under a burst of requests.}
    \label{fig:response_dist}
\end{figure}

\begin{table*}[!t]
\caption{Comparison of performance on downstream tasks with full data size.
         We report accuracy for classification tasks and entity-based F1 for named entity recognition tasks.}
\scriptsize
\setlength{\tabcolsep}{2.2pt}
\begin{tabular}{@{}lccccccccccc@{}}
\toprule
\multirow{2}{*}{Method}       
             & Chemprot & BC5CDR & JNLPBA & NCBI-disease & Rct-20k & NewsGroup & ACL-ARC & SciERC & Citation Intent & Sci-Cite & \multirow{2}{*}{Avg.} \\
             & Bio/\textsc{CLS} 
                        & Bio/\textsc{NER} 
                                 & Bio/\textsc{NER} 
                                          & Bio/\textsc{NER}      & Med/\textsc{CLS} & Eco/\textsc{CLS}   & CS/\textsc{CLS}  & CS/\textsc{NER} & Sci/\textsc{CLS} & Sci/\textsc{CLS} &  \\
\midrule
\textsf{BERT-dedicated}          & \textbf{0.8242} & \underline{0.8533} & \underline{0.7184} & \textbf{0.8439} & \textbf{0.8689} & \textbf{0.8775} & \textbf{0.7266} & \underline{0.6475} & 0.8058 & \underline{0.8705} &  \underline{0.8034}   \\
\textsf{BERT-shared}             & 0.3603 & 0.6399 & 0.4185 & 0.5680 & 0.5237 & 0.4896 & 0.3525 & 0.4404 & 0.5180 & 0.5341 &  0.4845   \\
\textsf{BERT-compress}\textsubscript{distil}              & 0.5520 & 0.5815 & 0.5726 & 0.4657 & 0.8178 & 0.7420 & 0.4892 & 0.2633 & 0.6331 & 0.7818 & 0.5899 \\
\textsf{BERT-compress}\textsubscript{tiny6}     
                        & 0.7899 & 0.8422 & 0.7124 & 0.8153 & 0.8642 & 0.8623 & 0.6331 & 0.6260 & 0.7914 & 0.8550 & 0.7792 \\
\textsf{BERT-compress}\textsubscript{tiny4} 
                        & 0.6993 & 0.8001 & 0.6949 & 0.7901 & 0.8440 & 0.8404 & 0.6115 & 0.5486 & 0.7626 & 0.8270 & 0.7419 \\
\hline
\textsf{hBERT-adapter}      
                        & 0.7723 & 0.8283 & 0.7137 & 0.8368 & 0.8640 & 0.8655 & 0.6475 & 0.6169 & 0.7410 & 0.8512 &   0.7737  \\
\textsf{hBERT-PLOT} 
                        & 0.8146 & \textbf{0.8568} & \textbf{0.7200} 
                                                 & 0.8400 & \underline{0.8685}  & \underline{0.8695} & \underline{0.7194} & \textbf{0.6489} 
                                                                                                & \textbf{0.8345} 
                                                                                                         & \textbf{0.8716} 
                                                                                                                  &  \textbf{0.8043}  \\
\textsf{hBERT}                   & \underline{0.8161} & 0.8493 & 0.7173 & \underline{0.8437} &  0.8634  & 0.8674 & 0.6547 & 0.6323 & \underline{0.8201} & 0.8651 & 0.7929  \\
\bottomrule
\end{tabular}
\label{tab:accuracy}
\end{table*}

\paragraph{End-to-end Response Time.}
The \texttt{infer} requests will first be inserted into an input queue when received. 
Our system continuously monitors the queue, and processes the requests in batches, whenever the queue is non-empty. 
The end-to-end response time is then defined as the duration between the time a request enters the queue and the time it is processed and output.
We here assume each tenant sends \texttt{infer} requests at a constant rate, i.e.,~one request for every 20s,
and each tenant sends 100 requests in this test.

\cref{fig:response_time} shows the results of response time with a different number of tenants.
When the number of tenants is small, since the batch size is mostly one at the server side, the response time remains constant.
As the number of tenants increases, the response time increases significantly for all the baseline methods.
We can observe that \textsf{hBERT} still maintains a reasonable response time even when 10K tenants keep sending \texttt{infer} requests.
In the meantime, other baselines were unable to process the requests in time.
The results reveal the effectiveness of \textsf{hBERT} in handling a large number of \texttt{infer} requests from multiple tenants, while maintaining a low end-to-end response time.

\begin{figure}[!t]
\centering
\includegraphics[width=1\linewidth]{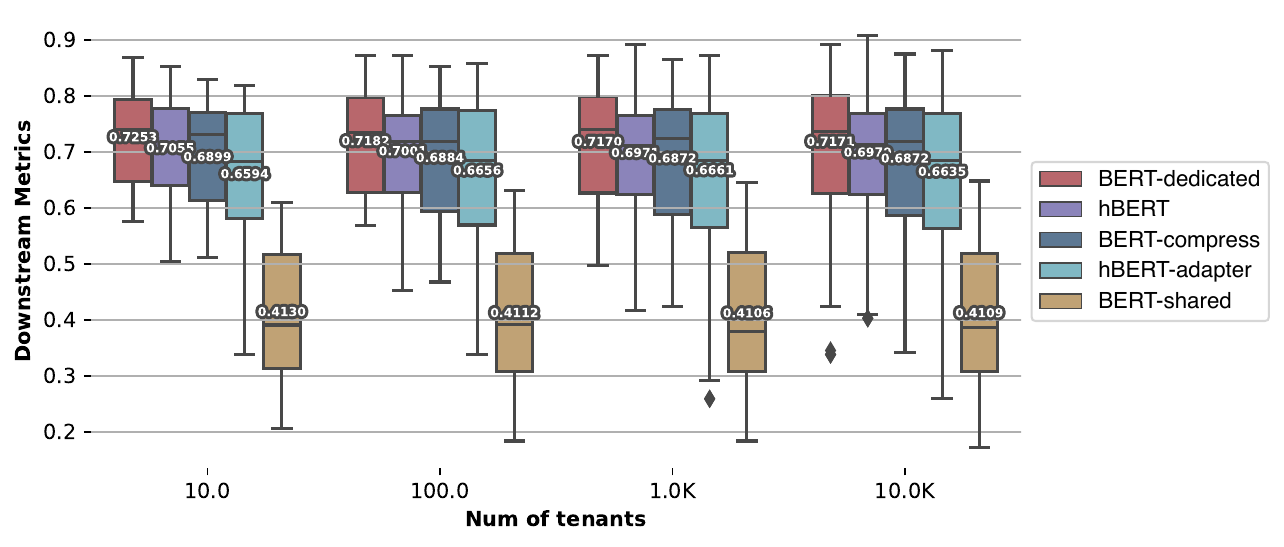}
\caption{Downstream metrics with different number of tenants. 
          We annotate the average metric of classification tasks (accuracy) and named entity recognition tasks (F1).
         }
\label{fig:metric}
\end{figure}

\begin{figure}[t]
    \centering
    \includegraphics[width=1.0\linewidth]{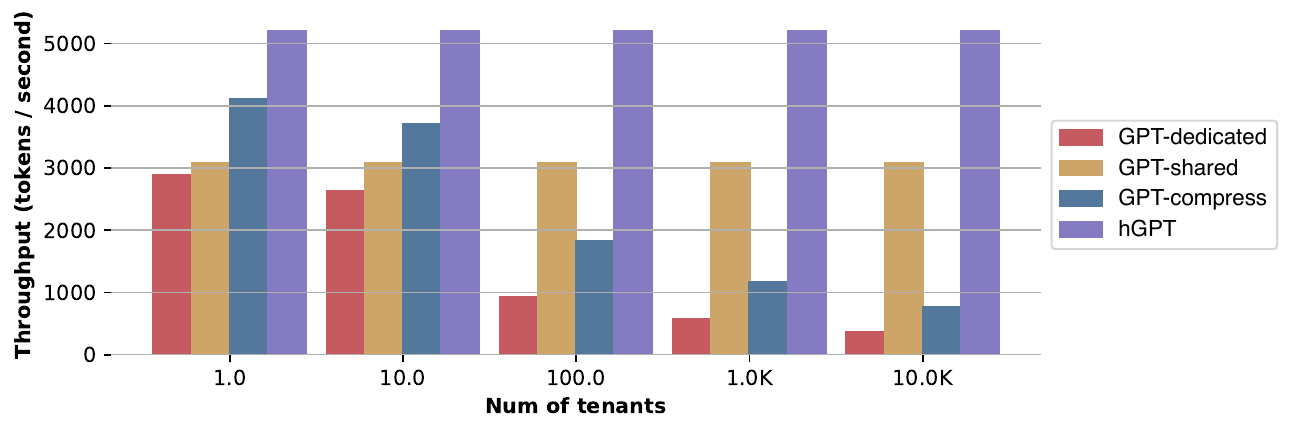}
    \caption{Token throughput (the number of generated tokens per second) of generative PLMs with different \#tenants.}
    \label{fig:throughput_gpt}
\end{figure}


\paragraph{Burst Requests.}
Burstiness is an inherent characteristic of real-world workloads \cite{xue2016managing}.
To better understand the performance of our approach in this scenario,
we simulate bursty workloads by sending \texttt{infer} requests in a short time and then compare the cumulative distribution of response time in \cref{fig:response_dist}.
Specifically, we launch 250 threads at intervals of 0.05s.
Each thread sends an \texttt{infer} request every 1.2s, and stops after sending 50 requests,
which leads to 12.5K requests in 72.5s.
By comparing with other approaches, the overall response distribution of \textsf{hBERT} is left-shifted and thus yields a smaller average response time. 

\subsection{Results on Downstream Tasks} \label{exp:acc}
In this subsection, we compare the accuracy of different methods on multiple downstream tasks.

\cref{tab:accuracy} presents the results using the full data of each dataset.
\textsf{BERT-dedicated} could be considered as an upper bound in terms of the accuracy.
However, it is not suitable for serving hundreds of tenants with limited hardware resources.
Although \textsf{BERT-shared} does not introduce any task-specific parameters,
the accuracy degrades on all tasks.
\textsf{BERT-compress} uses knowledge distillation to reduce the model size, and it also hurts the accuracy.
Moreover, it cannot address the `massive instances inference' problem.
They can only achieve that by model swapping which leads to poor runtime performance.
In comparison, our approach achieves the best average accuracy except \textsf{BERT-dedicated},
showing the effectiveness of our approach in terms of downstream accuracy.
\cref{fig:metric} presents the results with different number of tenants.
We randomly sample a subset of each dataset to simulate 10, 100, 1K, 10K tenants.  
And the result shows that our approach outperforms other schemes consistently. 
 \begin{table*}[!t]
\caption{The performance of text summarization and text rewriting.}
\centering
\setlength{\tabcolsep}{2pt} 
\renewcommand{\arraystretch}{1.2} 
\resizebox{\textwidth}{!}{ 
\begin{tabular}{@{}l c | c c c | c c c | c c c@{}}
\toprule
\multirow{2}{*}{Method} & \multirow{2}{*}{Latency} & \multicolumn{3}{c}{CNN/DailyMail} & \multicolumn{3}{|c}{XSUM} & \multicolumn{3}{|c}{Quora}\\
\cmidrule(lr){3-11}
 &  & ROUGE-1 & ROUGE-2 & ROUGE-L & ROUGE-1 & ROUGE-2 & ROUGE-L & ROUGE-1 & ROUGE-2 & ROUGE-L\\
\midrule
\textsf{GPT-compress}\textsubscript{distil}  & -49\% & 0.3209 & 0.1243 & 0.3037 & 0.2709 & 0.0816 & 0.2220  & 0.3578 & 0.1856 & 0.3458\\
\textsf{GPT-compress}\textsubscript{tiny6} & -49\% & 0.3199 & 0.1248 & 0.3032 & 0.2751 & 0.0840 & 0.2264 & 0.3547 & 0.1812 & 0.3431\\
\textsf{hGPT-PLOT}\textsubscript{6+6} &  -49\% & 0.3283 & 0.1281 & 0.3098 & 0.2819 & 0.0914 & 0.2319 &0.3587 & 0.1863 & 0.3472 \\
\bottomrule
\end{tabular}
}
\label{tab:extend_GPT}
\end{table*}

\begin{figure}[t]
    \includegraphics[width=\linewidth]{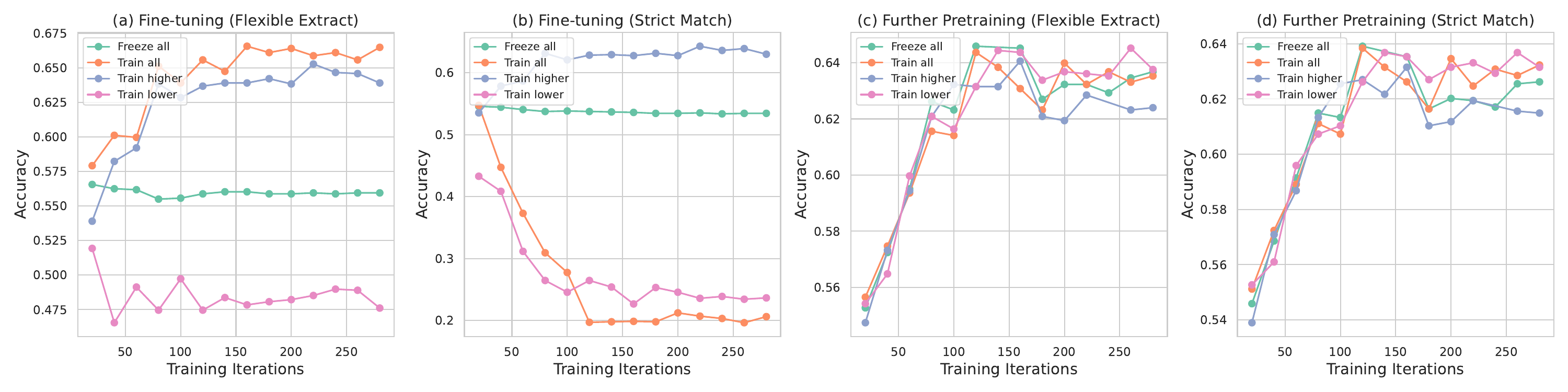}
    \caption{\where{Verification of Hierarchical Knowledge in the Decoder-Only Large Language Model (LLaMA-3.1-8B): Fine-Tuning with LoRA on GSM8K Training Set and Further Pretraining on OpenWebMath. Evaluated using CoT and 8-shot on the GSM8K Test Set, Reporting Two Answer Filtering Patterns: Flexible Extract and Strict Match.}}
    \label{fig:hier_llama}
\end{figure}

\begin{table}[t]
    \caption{Effect of domain-incremental space usage to the accuracy.}
    \centering
    \small
    \setlength{\tabcolsep}{3pt}
    \begin{tabular}{@{}lcrcrcr@{}}
    \toprule
    \multirow{2}{*}{Task} & \multicolumn{2}{c}{BC5CDR}   & \multicolumn{2}{c}{NewsGroup}    &   \multicolumn{2}{c}{Chemprot}    \\
                  & \multicolumn{2}{c}{Bio/\textsc{NER}}  & \multicolumn{2}{c}{Eco/\textsc{CLS}}         &   \multicolumn{2}{c}{Bio/\textsc{CLS}}    \\ 
    \cmidrule(lr){2-3} \cmidrule(lr){4-5} \cmidrule(lr){6-7}
    Metric          &  F1    &  Space           &  Acc      &  Space              &  Acc           & Space            \\ 
    \midrule
    $\alpha=100\%$  & 0.8480 &  463GB    & 0.8543 &  339GB  & 0.8025  & 463GB               \\
    $\alpha=50\%$   & 0.8458 &  1.42GB   & 0.8731 & 1.75GB  & 0.7979  & 1.42GB               \\
    $\alpha=30\%$   & 0.8431 &  0.31GB   & 0.8680 & 0.13GB  & 0.7942  & 0.31GB               \\
    $\alpha=0\%$    & 0.8417 &  -        & 0.8674 &    -    & 0.7916  &  -             \\
    \bottomrule
    \end{tabular}
    \label{tab:space_acc}
\end{table}

\subsection{Extension to Generative PLMs}
\label{sec:GPT}
In this subsection, we demonstrate that our scheme can adopt other PLMs, in particular
the widely used GPT \cite{gpt} variants.
Since GPT models are unable to access future tokens during inference, our scheme precomputes text fragments, denoted as $\v X_i = [x_{i-2}, x_{i-1}, x_{i}]$, to enable efficient lower layer skipping.
These fragments are stored in PLOT for quick retrieval during inference.

We use OPT-125M \cite{zhang2022opt} as the backbone model.
To evaluate the effectiveness of our approach, we compare the token throughput of our adapted scheme \textsf{hGPT}, 
with different schemes which are likewise named \textsf{GPT-dedicated}, \textsf{GPT-shared}, and \textsf{GPT-compress} (using \textsf{DistilGPT2} \cite{sanh2019distilbert} due to the absence of compressed OPT models). 
\cref{fig:throughput_gpt} shows that \textsf{hGPT} achieves a consistently high token throughput with 1 to 10K tenants, 
making it a cost-effective solution for large-scale deployment.

\changeone{For generation performance, \rone{R1.\\C22}we conducted\rthree{R3.\\C1} additional experiments on \textsf{hGPT-PLOT}\textsubscript{6+6}, where the PLOT was constructed using the bottom six layers of \textsf{hGPT} trained on the Pile corpus~\cite{gao2020pile}. The experiments include fine-tuning the six-layer \textsf{hGPT}, \textsf{GPT-compress}\textsubscript{distil}, and \textsf{GPT-compress}\textsubscript{tiny6} models on the training sets of the CNN/DailyMail and XSum~\cite{Narayan2018DontGM} text summarization tasks, as well as the Quora~\cite{chandra2020Quora} text rewriting task. We measure the generation performance of different baselines (corresponding models are all GPU-resident) in multiple natural language generation tasks while achieving the same inference latency. As shown in~\cref{tab:extend_GPT}, \textsf{hGPT-PLOT}\textsubscript{6+6} achieve ROUGE-1 scores of 0.3283 and 0.2819 on CNN/DailyMail and XSum text summarization, respectively, and in the text rewriting task (Quora). A ROUGE-1 score of 0.3587 was achieved, which outperforms the compressed baselines in performance.}

\changeone{We acknowledge\rtwo{R2.\\C3,C4} the necessity of longer context lengths for large models, a widely validated insight supported by prior works, especially those focused on key-value compression of LLM. These studies~\cite{XiaoTCHL24,zhuo2024magicpig} emphasize the importance of both the context window (essentially our $n$-gram) and certain attention sinks. 
Although models with billions of parameters are not ideally suited for our setting, as they may not be feasible for multi-tenant inference on resource-constrained GPUs, the hierarchical knowledge distribution characteristic remains prevalent in large models.} 

\changeone{To assess the generalizability of hierarchical\rtwo{R2.\\C1} knowledge, we extend our analysis to decoder-only models, conducting experiments on the GSM8K~\cite{cobbe2021gsm8k} mathematical problem-solving task using the larger LLaMA-3.1-8B model. As shown in~\cref{fig:hier_llama} (a, b), we observe that fine-tuning the higher layers (top 16 layers) of the model leads to strong performance in both flexible extraction and strict matching, consistent with our findings from encoder-only models.}

\changeone{Furthermore, fine-tuning the lower layers (Train all and Train lower) on task-specific data can degrade performance, especially in the strict matching case. We hypothesize that the lower layers of the model learn domain-specific knowledge relevant to mathematics, and fine-tuning on specific tasks disrupts the foundational processing patterns established in these layers.}

\changeone{In contrast, as depicted in~\cref{fig:hier_llama} (c, d), further pretraining on mathematics-specific corpora (OpenWebMath~\cite{paster2023openwebmath}) enables the lower layers to better capture domain-relevant processing patterns, leading to improved performance. This observation aligns with our findings in encoder-only models.
Therefore, replacing PLOT with sparse attention mechanisms~\cite{XiaoTCHL24,zhuo2024magicpig}, which offer a promising approach to reducing lower layer computation.  
Additionally, speculative decoding~\cite{Zhang00S0CM24} could be leveraged to ensure lossless acceleration.}  

\subsection{Additional Space Requirements} \label{exp:plot}
The HMI requires the cloud platform to store the materialized hidden states of lower transformer layers of \textsf{hPLM} in the form of PLOT, which are preserved at host memory and SSD.
For different domains of data, we conduct further pretraining and save the domain-specific hidden states to enhance the downstream accuracy.
We note that this incurs minor additional space usage -- 
for most text fragments, their representations do not differ much under different domains, so we can share them without duplicating them multiple times;
for text fragments whose distribution and semantics differ significantly between domains, we keep domain-specific copies of them.
Here, we show that storing the top $\sim$0.4\% frequent trigrams can cover the 50\% occurrences in the domain-specific pretraining corpus, and can achieve almost the same accuracy as, sometimes better than, storing all trigrams for a new domain.

\cref{tab:space_acc} shows the accuracy and space requirements by preserving top $\alpha\%$ occurrences of new domain trigrams for \textsf{hBERT}.
$\alpha=100\%$ means all domain-specific trigrams are preserved; $\alpha=0\%$ means only using the root version of PLOT.
We surprisingly find that a moderate threshold $\alpha=50\%$ sometimes achieves a better accuracy than $\alpha=100\%$, which shows that frequent trigrams contribute the most to the downstream task,
while others are trivial to the result and thus can be pruned.
Thus, we only need to store several GBs for each domain, an acceptable size for host memory.
Following the same procedure, we process data for a total of 10 domains, requiring only 18.9 GB of additional storage.
Notably, the cost of host memory and SSD is extremely low compared to VRAM.

\subsection{Effect of Pipeline Optimization} \label{exp:pipeline} 

\begin{table}[t]
    \caption{Effect of pipeline optimization.}
    \centering
    \small
    \setlength{\tabcolsep}{1pt}
    \begin{tabular}{@{}ll@{}}
    \toprule
        Setting  & Throughput (16 tenants) \\
    \midrule
        \textsf{BERT-dedicated} 
                 & 165.2 req/s \\
        \textsf{hBERT}    & \textbf{380.9 req/s} (2.31x) \\
        \quad w/o pipelined retrieval
                 & 340.1 req/s \  (2.06x) \\
        \quad w/o pipelined adapter swapping
                 & 338.9 req/s \  (2.05x) \\
        \quad synchronized
                 & 304.8 req/s \  (1.85x) \\
    \bottomrule
    \end{tabular}
    \label{tab:pipeline}
\end{table}
\begin{figure}[!t]
    \centering
    \includegraphics[width=1.0\linewidth]{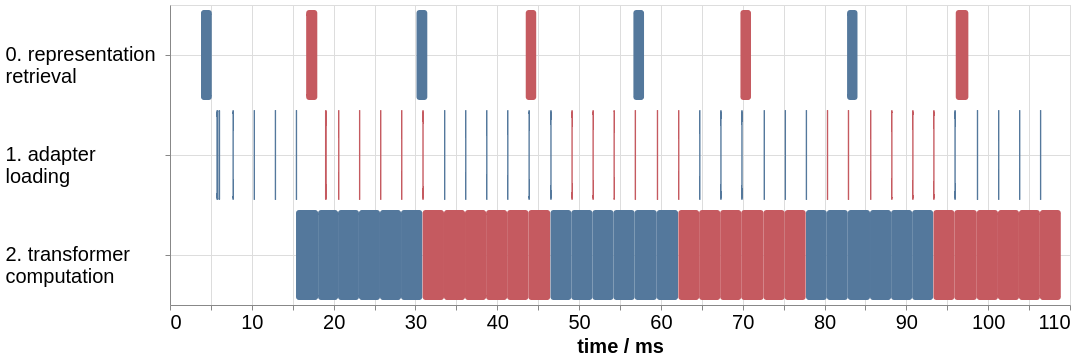}
    \caption{Breakdown of the inference pipeline. Consecutive inference batches are colored differently.}
    \label{fig:breakdown}
\end{figure}
\begin{table*}[!t]
\caption{The performance of a set of general NLP
understanding tasks.}
\centering
\setlength{\tabcolsep}{1.5pt}
\centering
\begin{tabular}{@{}l c c c c c c c c c c c @{}}
\toprule
{Method} & {Latency} & CoLA & SST-2 & MRPC & STS-B & QQP & MNLI & QNLI & RTE & WNLI & SQuAD\\
\midrule
\textsf{BERT-compress}\textsubscript{distil6}  & -49\% & 49.0 & 92.5 & 86.9 & -/81.3 & 70.1/- & 82.6/81.3  & 88.9 & 58.4 & - & 78.1/86.2\\
\textsf{BERT-compress}\textsubscript{tiny6} & -49\% & 46.1 & 92.6 & 88.0/- & -/83.9 & 71.3/- & 84.4/83.1 & 89.8 & 69.7 & - & 81.6/88.8\\
\textsf{hBERT-PLOT}\textsubscript{6+6} &  -49\% & 52.7 & 93.3 & 88.9/85.0 & 87.0/85.8 & 71.9/89.2 & 84.3/84.2 & 90.6 & 70.6 &  65.1 & 82.1/89.1 \\
\midrule
\textsf{BERT-compress}\textsubscript{distil4}  & -66\% & 32.8 & 91.4 & 82.4/- & -/76.1 & 68.5/- & 78.9/78.0  & 85.2 & 54.1 & - & 71.8/81.2\\
\textsf{BERT-compress}\textsubscript{tiny4} & -66\% & 25.3 & 90.0 & 85.4/- & -/80.4 & 68.9/- & 81.2/80.3 & 86.2 & 63.9 & - & 73.9/82.6\\
\textsf{hBERT-PLOT}\textsubscript{6+4} &  -66\% & 39.8 & 91.3 & 87.7/82.7 & 84.1/82.8 & 70.4/88.3 & 82.0/81.6 & 88.5 & 66.1 &  65.1 & 76.2/84.5 \\
\bottomrule
\end{tabular}
\label{tab:extend_BERT}
\end{table*}
In this subsection, we conduct an ablation study on pipeline optimization.
\cref{tab:pipeline} shows the throughput for different settings.
Pipelined representation retrieval hides the time of searching hidden states in a previous batch of GPU computation.
Pipelined adapter swapping hides the time of loading adapter in a previous layer of GPU computation.
We can see disabling either pipelined representation retrieval or pipelined adapter swapping will cause a slowdown.
When we remove both pipeline optimization techniques, i.e.,~synchronized execution, 
while still faster than \textsf{BERT-shared}, it is about 20\% slower than our optimized \textsf{hBERT}.
This shows that our parallel implementation of HMI is highly optimized and demonstrates the importance of pipeline optimization for inference efficiency.

\paragraph{Breakdown of Inference Pipeline.} 
To better understand how the pipeline strategy improves inference efficiency, we here show the breakdown of our inference pipeline.
\cref{fig:breakdown} visualizes the profiling result of runs with a batch size of 4, each consisting of \texttt{infer} requests from 1,000 \textsf{hBERT} instances.
We can have several observations from the result:
1) The time to search the precomputed representations is much smaller than GPU computations, 
so we can overlap them and hide searching representations within GPU computations;
2) The time of loading adapters is also much smaller than GPU computations,  
so with the help of fine-grained pipeline parallelism, adapters for a particular transformer layer can be loaded after the previous batch of computations for that transformer layer and before the current batch of computations.
So we do not introduce additional inference time when searching the precomputed representations and loading adapter modules to the GPU device.
In general, with our efficient inference pipeline design,
the overall throughput will not slow down by additional CPU and I/O operations, 
which can be hidden within the GPU compute time.

\subsection{Evaluation of \textsf{hBERT} on Diverse Tasks} \label{exp:more_acc_bert}
\changeone{We further explore diverse natural\rthree{R3.\\C1} language understanding tasks using \textsf{hBERT}. Specifically, we construct \textsf{hBERT-PLOT}\textsubscript{6+6} and \textsf{hBERT-PLOT}\textsubscript{6+4} (following~\cite{wang2022skipbert}), where the lower layer PLOT consists of 6 layers, and the upper layer comprises either 6 or 4 layers, respectively. These models are compared against baselines, including \textsf{BERT-compress}.}

\changeone{As shown in~\cref{tab:extend_BERT}, \textsf{hBERT-PLOT} achieves comparable latency to the baselines across different upper layer configurations, while demonstrating superior performance on most tasks, including the reading comprehension task SQuAD. This underscores the strong generalization capabilities of \textsf{hBERT} in natural language understanding tasks.}

\begin{table}[t]
\caption{Impact of $n$-grams on \textsf{hBERT}.}
\centering
\setlength{\tabcolsep}{6pt}
\begin{tabular}{@{}l c c c@{}}
\toprule
\textsf{hBERT} Variants & CoLA & MRPC & MNLI \\
\midrule
\textsf{hBERT-PLOT}\textsubscript{2-gram} & 29.1 & 86.3/90.1 & 80.8/81.0 \\
\textsf{hBERT-PLOT}\textsubscript{3-gram} & 32.9 & 86.3/90.4 & 80.9/81.1 \\
\textsf{hBERT-PLOT}\textsubscript{5-gram} & \textbf{34.5} & 86.8/90.7 & \textbf{81.1/81.3} \\
\bottomrule
\end{tabular}
\label{tab:ngram_bert}
\end{table}

\begin{table}[t]
\caption{Impact of $n$-grams on \textsf{hGPT}.}
\centering
\setlength{\tabcolsep}{9pt}
\begin{tabular}{@{}l c c@{}}
\toprule
\textsf{hGPT} Variants & Pile-PPL@1K & Pile-PPL@5K \\
\midrule
\textsf{hGPT-PLOT}\textsubscript{2-gram} & 30.978 & 17.010 \\
\textsf{hGPT-PLOT}\textsubscript{3-gram} & 22.048 & 13.845 \\
\textsf{hGPT-PLOT}\textsubscript{5-gram} & \textbf{19.128} & \textbf{13.451} \\
\bottomrule
\end{tabular}
\label{tab:ngram_gpt}
\end{table}

\subsection{Design and Impact of PLOT Configurations} 

\changeone{The effectiveness of PLOT \rone{R1.\\C13}\rtwo{R2.\\C1,C3}\rthree{R3.\\C2}in HMI is theoretically supported by principles from prior research on transformer architectures~\cite{jawahar2019does,rogers2020primer,wang2022skipbert}. Lower transformer layers primarily capture short-range contexts, while upper layers model long-range dependencies. Consequently, token representations in lower layers remain stable, enabling the masking of distant tokens without losing critical information.}

\changeone{We also analyze how different configurations, including context window size ($n$-grams) and the number of lower layers (\# lower layers) affect PLOT's performance. An appropriately balanced configuration offers the most cost-effective solution.}

\paragraph{Impact of $n$-grams}
\changeone{We evaluated the effect of $n$-grams while fixing both lower and upper layers to 6:}
\begin{itemize}
    \item \textsf{hBERT} Results:  
    \changeone{As shown in \cref{tab:ngram_bert}, increasing $n$ improves performance by capturing larger local contexts. However, for $n > 3$, there is a combinatorial increase in storage and computation costs. For efficiency, we adopt a default 3-gram size.}
    
    \item \textsf{hGPT} Results: 
    \changeone{Similar trends are observed in \cref{tab:ngram_gpt}, where performance improves with increasing $n$. However, memory usage rises significantly for $n > 3$, leading us to set the default context size to 3-grams for \textsf{hGPT}.}
\end{itemize}

\paragraph{Impact of \# Lower Layers}
\changeone{We analyzed the effect of varying the number of lower layers while fixing the 3-grams for PLOT:}
\begin{itemize}
    \item \textsf{hBERT} Results:
    \changeone{As shown in \cref{tab:nlayer_bert}, increasing the number of lower layers enhances performance on tasks like CoLA, MRPC, and MNLI. However, excessive layers result in diminishing returns, as overfitting on small context windows reduces generalization. Based on these findings, the default depth for \textsf{hBERT} is set to 6 layers.}
    
    \item \textsf{hGPT} Results:  \changeone{\cref{tab:nlayer_gpt} shows a similar trend for perplexity (PPL) improvement as local depth increases. However, deeper layers may overfit due to limited short-range context processing. We therefore select 6 layers as the default for \textsf{hGPT} to balance depth and generalization.}
\end{itemize}

\begin{table}[t]
\caption{Impact of \# lower layer on \textsf{hBERT}.}
\centering
\setlength{\tabcolsep}{7pt}
\begin{tabular}{@{}l c c c@{}}
\toprule
\textsf{hBERT} Variants & CoLA & MRPC & MNLI \\
\midrule
\textsf{hBERT-PLOT}\textsubscript{2+4} & 28.6 & 86.0/90.0 & 80.5/81.0 \\
\textsf{hBERT-PLOT}\textsubscript{4+4} & 31.4 & 86.0/90.2 & 80.9/81.1 \\
\textsf{hBERT-PLOT}\textsubscript{6+4} & \textbf{32.9} & \textbf{86.3/90.4} & 80.9/81.1 \\
\textsf{hBERT-PLOT}\textsubscript{8+4} & 32.4 & 85.0/89.5 & \textbf{81.0/81.2} \\
\bottomrule
\end{tabular}
\label{tab:nlayer_bert}
\end{table}
\begin{table}[t]
\caption{Impact of \# lower layer on \textsf{hGPT}.}
\centering
\setlength{\tabcolsep}{11pt}
\begin{tabular}{@{}l c c@{}}
\toprule
\textsf{hGPT} Variants & Pile-PPL@1K & Pile-PPL@5K \\
\midrule
\textsf{hGPT-PLOT}\textsubscript{2+6} & 24.395 & 15.040 \\
\textsf{hGPT-PLOT}\textsubscript{4+6} & 23.188 & 13.935 \\
\textsf{hGPT-PLOT}\textsubscript{6+6} & \textbf{23.546} & 13.845 \\
\textsf{hGPT-PLOT}\textsubscript{8+6} & 23.864 & \textbf{13.839} \\
\bottomrule
\end{tabular}
\label{tab:nlayer_gpt}
\end{table}

\section{Related Work} \label{sec:related}

\paragraph{Pretrained Language Models.}
With the prevalence of PLM \cite{bert,gpt,albert,roberta,xlnet}, in NLP applications,
people tend to choose pretrained models and fine-tune them on customized datasets.
PLMs are usually based on transformer \cite{vaswani2017attention}, consisting of a self-attention sub-layer and a feedforward sub-layer.
The self-attention layer takes as input a sequence of token representations, 
learns the relationship between every two tokens as the attention weights, and produces a new sequence of representations.
During the pretraining phase, a PLM is trained using an extensive upstream corpus,
usually in a self-supervised manner, e.g.,~causal language modeling \cite{gpt} and masked language modeling \cite{bert}.
After obtaining PLM, it is also recommended to further pretrain it using domain-specific corpus \cite{jin2021lifelong} to enhance the domain knowledge and get domain-specific PLMs.

During the fine-tuning phase, we shall start from a PLM and refine its parameters for multiple iterations using a gradient descent-based optimizer.
Despite the good downstream accuracy of this strategy,
it changes all parameters and will produce a distinct model for each downstream task, which is space-inefficient.
To tackle this problem, parameter-efficient fine-tuning, e.g., Adapter \cite{houlsby2019parameter}, is proposed: 
instead of refining all parameters of PLMs, one can fine-tune or insert only a small number of parameters while achieving similar performance to fully fine-tuning.
This approach allows us to deploy models for multiple tasks in sub-linear space complexity. 
LoRA \cite{hu2021lora} freezes the pre-trained model weights and injects trainable rank decomposition matrices.
Prefix-tuning \cite{li2021prefix} adds ``virtual tokens'' as a prefix and only updates the prefix.
Prompt tuning \cite{lester2021power} learns soft prompts rather than hand-crafted prompts.
Study \cite{he2021towards} provides a unified vision of parameter-efficient fine-tuning.
However, although they work well for adapting PLMs to small-scale data, 
it usually under-perform with large training data, or the downstream data is significantly different from the upstream data in pretraining.

\paragraph{Model Compression.}
One of the limiting factors hampering the adoption of PLMs
is the large number of parameters that these models have, 
which leads to a large GPU memory footprint and increased inference time. 
Knowledge distillation \cite{hinton2015distilling}
provides an effective way to transfer the knowledge embedded in a teacher network to a student network.
The student network is usually more lightweight than the teacher network and thus more computationally efficient.
The student network can be structurally identical to the teacher but contains
fewer layers or hidden units, 
e.g.,~BERT-PKD \cite{sun2019bertpkd}, DistilBERT \cite{sanh2019distilbert}, TinyBERT \cite{jiao2020tinybert}, MiniLM \cite{wang2020minilm}, and BERT-EMD \cite{bertemd}.
Meanwhile, 
some work adopts specifically designed networks, e.g.,~SqueezeBERT \cite{squeezebert} and MobileBERT \cite{mobilebert}, to reduce the computation per layer.
SkipBERT \cite{wang2022skipbert} proposes to precompute the shallow hidden states to accelerate the inference latency, but they only consider the single-model version case without delving into shallow feature version management.
Quantization is another way to reduce the model volume and accelerate the inference speed on specific hardware.
Application on PLMs include Q8BERT \cite{zafrir2019q8bert}, BinaryBERT \cite{bai2021binarybert}, I-BERT \cite{kim2021bert}, BiBERT \cite{qin2021bibert}.
In general, model compression reduces the number of parameters / volume of the model
at the cost of accuracy, 
however, model compression usually harms the downstream accuracy, and it is still challenging to accommodate hundreds and even thousands of models in one single GPU.
Besides, quantization methods, usually with lower bits, often require specialized hardware to observe end-to-end speedup.

\paragraph{Model Deployment.}
Studies~\cite{olston2017tensorflow,chard2019dlhub,zhang2019mark} have been developed to simplify the model deployment process. 
However, most of these systems consider model instances as black boxes, 
with the focus being primarily on ease of deployment rather than inference cost.
To optimize model inference, these systems use model-agnostic techniques such as batching, buffering, and caching \cite{crankshaw2017clipper}.
The concept of white box model serving \cite{lee2018pretzel} has been introduced, enabling model-specific optimizations through resource sharing and parameter reuse. Nonetheless, these optimizations primarily target small-sized models and non-neural operations like tokenization and feature concatenation.
Recently, PetS \cite{zhou2022pets} offered a unified framework for multi-task PETs serving. 
S-LoRA \cite{sheng2023s} was developed to serve a large number of LoRA models on a single GPU. \cite{chen2023punica} introduced a CUDA kernel design for batch inference of low-rank matrix multiplication. 
While these works primarily focus on task-specific models with lossless accuracy, our work aims to address both domain-specific and task-specific models with minor accuracy degradation.

\paragraph{Model Management.}
In addition to the numerous ML inference serving systems, 
the data management community has also been exploring model management, 
resulting in several noteworthy works such as Cerebro \cite{nakandala2020cerebro} and ModelDB \cite{vartak2016modeldb} and follow-up work \cite{kumar2017data,li2021intermittent,li2021towards,schelter2015challenges}. 
These works enable the efficient management of ML models by providing end-to-end support for various stages of the model lifecycle, such as training, deployment, and monitoring.
There exist pre-trained model repositories, most notably Tensorflow Hub, PyTorch Hub, and HuggingFace\footnote{
\url{https://tfhub.dev},
\url{https://pytorch.org/hub},
\url{https://huggingface.co/models}
}.
Each repository has its own interface for accessing those models.
These repositories allow easy access to pretrained models along with their parameters and additional meta-data 
such as the domains and tasks that the models are designed for.
However, they mainly focus on model storage, usually on a hard disk or SSD,
and model searching, without considering their deployment and inference.
And this study presents a co-design of storage and computation scheme to manage and focus a massive number of PLM model instances.

\section{Conclusion} \label{sec:conclusion}
In this study, we presented an innovative prototype system capable of managing and serving a massive number of PLMs on a single commodity cloud server with one GPU. 
By constructing hierarchical PLMs (\textsf{hPLM}s) and thus batching inference requests,
our system provided a cost-effective and scalable solution for provisioning PLMs. 
We introduced specialized data structures that encapsulate the domain and task-specific knowledge learned by each PLM. 
We designed domain-specific knowledge trees based on frequency update strategy and task-specific knowledge swapping to hierarchically manage the knowledge in \textsf{hPLM}s generated by massive tenants with acceptable storage increases and limited GPU memory.
Additionally, we proposed several system optimization techniques to enhance inference throughput, further solidifying our system's efficiency and scalability.

Our empirical study substantiated the effectiveness of our proposed system, 
demonstrating its ability to simultaneously serve up to 10,000 model instances from a single GPU (NVIDIA Quadro P5000 with 16GB VRAM).
Impressively, this was achieved while maintaining high throughput, low response time, and without sacrificing the model quality. 
We conducted extensive tests on both encoder-based and decoder-based PLMs to showcase the generality of our approach and its superiority compared to existing methods.

In the future, we plan to extend our system to accommodate more types of models, 
such as computer vision pretrained models. 
\changeone{\rthree{R3.\\C3}We will also investigate cost-effective training schemes, such as other parameter-efficient fine-tuning schemes (such as the popular adapter LoRA in decorder-only PLM should be pluggable for HMI),
further enhancing the versatility and efficiency of our system.}

\bibliographystyle{spmpsci}
\bibliography{unified.bib}

\begin{thebibliography}{10}
\providecommand{\url}[1]{{#1}}
\providecommand{\urlprefix}{URL }
\expandafter\ifx\csname urlstyle\endcsname\relax
  \providecommand{\doi}[1]{DOI~\discretionary{}{}{}#1}\else
  \providecommand{\doi}{DOI~\discretionary{}{}{}\begingroup \urlstyle{rm}\Url}\fi

\bibitem{baek2020multi}
Baek, E., Kwon, D., Kim, J.: A multi-neural network acceleration architecture.
\newblock In: 2020 ACM/IEEE 47th Annual International Symposium on Computer Architecture (ISCA), pp. 940--953. IEEE (2020)

\bibitem{bai2021binarybert}
Bai, H., Zhang, W., Hou, L., Shang, L., Jin, J., Jiang, X., Liu, Q., Lyu, M., King, I.: {B}inary{BERT}: Pushing the limit of {BERT} quantization.
\newblock In: Proceedings of the 59th Annual Meeting of the Association for Computational Linguistics and the 11th International Joint Conference on Natural Language Processing (Volume 1: Long Papers), pp. 4334--4348. Association for Computational Linguistics, Online (2021).
\newblock \doi{10.18653/v1/2021.acl-long.334}.
\newblock \urlprefix\url{https://aclanthology.org/2021.acl-long.334}

\bibitem{qwen}
Bai, J., Bai, S., Chu, Y., \textit{et al.}: Qwen technical report.
\newblock arXiv preprint arXiv:2309.16609  (2023)

\bibitem{scibert}
Beltagy, I., Lo, K., Cohan, A.: {S}ci{BERT}: A pretrained language model for scientific text.
\newblock In: Proceedings of the 2019 Conference on Empirical Methods in Natural Language Processing and the 9th International Joint Conference on Natural Language Processing (EMNLP-IJCNLP), pp. 3615--3620. Association for Computational Linguistics, Hong Kong, China (2019).
\newblock \doi{10.18653/v1/D19-1371}.
\newblock \urlprefix\url{https://aclanthology.org/D19-1371}

\bibitem{bergsma2021generating}
Bergsma, S., Zeyl, T., Senderovich, A., Beck, J.C.: Generating complex, realistic cloud workloads using recurrent neural networks.
\newblock In: Proceedings of the ACM SIGOPS 28th Symposium on Operating Systems Principles, pp. 376--391 (2021)

\bibitem{bommasani2021opportunities}
Bommasani, R., Hudson, D.A., Adeli, E., Altman, R., Arora, S., von Arx, S., Bernstein, M.S., Bohg, J., Bosselut, A., Brunskill, E., et~al.: On the opportunities and risks of foundation models.
\newblock ArXiv preprint \textbf{abs/2108.07258} (2021).
\newblock \urlprefix\url{https://arxiv.org/abs/2108.07258}

\bibitem{brown2020language}
Brown, T.B., Mann, B., Ryder, N., Subbiah, M., Kaplan, J., Dhariwal, P., Neelakantan, A., Shyam, P., Sastry, G., Askell, A., Agarwal, S., Herbert{-}Voss, A., Krueger, G., Henighan, T., Child, R., Ramesh, A., Ziegler, D.M., Wu, J., Winter, C., Hesse, C., Chen, M., Sigler, E., Litwin, M., Gray, S., Chess, B., Clark, J., Berner, C., McCandlish, S., Radford, A., Sutskever, I., Amodei, D.: Language models are few-shot learners.
\newblock In: H.~Larochelle, M.~Ranzato, R.~Hadsell, M.~Balcan, H.~Lin (eds.) Advances in Neural Information Processing Systems 33: Annual Conference on Neural Information Processing Systems 2020, NeurIPS 2020, December 6-12, 2020, virtual (2020).
\newblock \urlprefix\url{https://proceedings.neurips.cc/paper/2020/hash/1457c0d6bfcb4967418bfb8ac142f64a-Abstract.html}

\bibitem{chandra2020Quora}
Chandra, A., Stefanus, R.: Experiments on paraphrase identification using quora question pairs dataset (2020).
\newblock \urlprefix\url{https://arxiv.org/abs/2006.02648}

\bibitem{chard2019dlhub}
Chard, R., Li, Z., Chard, K., Ward, L., Babuji, Y., Woodard, A., Tuecke, S., Blaiszik, B., Franklin, M.J., Foster, I.: Dlhub: Model and data serving for science.
\newblock In: 2019 IEEE International Parallel and Distributed Processing Symposium (IPDPS), pp. 283--292. IEEE (2019)

\bibitem{chen2023punica}
Chen, L., Ye, Z., Wu, Y., Zhuo, D., Ceze, L., Krishnamurthy, A.: Punica: Multi-tenant lora serving.
\newblock arXiv preprint arXiv:2310.18547  (2023)

\bibitem{zhuo2024magicpig}
Chen, Z., Sadhukhan, R., Ye, Z., Zhou, Y., Zhang, J., Nolte, N., Tian, Y., Douze, M., Bottou, L., Jia, Z., Chen, B.: Magicpig: {LSH} sampling for efficient {LLM} generation.
\newblock CoRR \textbf{abs/2410.16179} (2024).
\newblock \doi{10.48550/ARXIV.2410.16179}.
\newblock \urlprefix\url{https://doi.org/10.48550/arXiv.2410.16179}

\bibitem{cobbe2021gsm8k}
Cobbe, K., Kosaraju, V., Bavarian, M., Chen, M., Jun, H., Kaiser, L., Plappert, M., Tworek, J., Hilton, J., Nakano, R., Hesse, C., Schulman, J.: Training verifiers to solve math word problems.
\newblock arXiv preprint arXiv:2110.14168  (2021)

\bibitem{crankshaw2017clipper}
Crankshaw, D., Wang, X., Zhou, G., Franklin, M.J., Gonzalez, J.E., Stoica, I.: Clipper: A $\{$Low-Latency$\}$ online prediction serving system.
\newblock In: 14th USENIX Symposium on Networked Systems Design and Implementation (NSDI 17), pp. 613--627 (2017)

\bibitem{bert}
Devlin, J., Chang, M.W., Lee, K., Toutanova, K.: {BERT}: Pre-training of deep bidirectional transformers for language understanding.
\newblock In: Proceedings of the 2019 Conference of the North {A}merican Chapter of the Association for Computational Linguistics: Human Language Technologies, Volume 1 (Long and Short Papers), pp. 4171--4186. Association for Computational Linguistics, Minneapolis, Minnesota (2019).
\newblock \doi{10.18653/v1/N19-1423}.
\newblock \urlprefix\url{https://aclanthology.org/N19-1423}

\bibitem{gao2020pile}
Gao, L., Biderman, S., Black, S., Golding, L., Hoppe, T., Foster, C., Phang, J., He, H., Thite, A., Nabeshima, N., Presser, S., Leahy, C.: The pile: An 800gb dataset of diverse text for language modeling (2020).
\newblock \urlprefix\url{https://arxiv.org/abs/2101.00027}

\bibitem{han2015deep}
Han, S., Mao, H., Dally, W.J.: Deep compression: Compressing deep neural networks with pruning, trained quantization and huffman coding.
\newblock In: International Conference on Learning Representations (2016)

\bibitem{he2021towards}
He, J., Zhou, C., Ma, X., Berg-Kirkpatrick, T., Neubig, G.: Towards a unified view of parameter-efficient transfer learning.
\newblock In: International Conference on Learning Representations (2021)

\bibitem{hinton2015distilling}
Hinton, G., Vinyals, O., Dean, J.: Distilling the knowledge in a neural network.
\newblock ArXiv preprint \textbf{abs/1503.02531} (2015).
\newblock \urlprefix\url{https://arxiv.org/abs/1503.02531}

\bibitem{houlsby2019parameter}
Houlsby, N., Giurgiu, A., Jastrzebski, S., Morrone, B., de~Laroussilhe, Q., Gesmundo, A., Attariyan, M., Gelly, S.: Parameter-efficient transfer learning for {NLP}.
\newblock In: K.~Chaudhuri, R.~Salakhutdinov (eds.) Proceedings of the 36th International Conference on Machine Learning, {ICML} 2019, 9-15 June 2019, Long Beach, California, {USA}, \emph{Proceedings of Machine Learning Research}, vol.~97, pp. 2790--2799. {PMLR} (2019).
\newblock \urlprefix\url{http://proceedings.mlr.press/v97/houlsby19a.html}

\bibitem{howard2018universal}
Howard, J., Ruder, S.: Universal language model fine-tuning for text classification.
\newblock In: Proceedings of the 56th Annual Meeting of the Association for Computational Linguistics (Volume 1: Long Papers), pp. 328--339. Association for Computational Linguistics, Melbourne, Australia (2018).
\newblock \doi{10.18653/v1/P18-1031}.
\newblock \urlprefix\url{https://aclanthology.org/P18-1031}

\bibitem{hu2021lora}
Hu, E.J., Wallis, P., Allen-Zhu, Z., Li, Y., Wang, S., Wang, L., Chen, W., et~al.: Lora: Low-rank adaptation of large language models.
\newblock In: International Conference on Learning Representations (2021)

\bibitem{huang2019clinicalbert}
Huang, K., Altosaar, J., Ranganath, R.: Clinicalbert: Modeling clinical notes and predicting hospital readmission.
\newblock ArXiv preprint \textbf{abs/1904.05342} (2019).
\newblock \urlprefix\url{https://arxiv.org/abs/1904.05342}

\bibitem{squeezebert}
Iandola, F., Shaw, A., Krishna, R., Keutzer, K.: {S}queeze{BERT}: What can computer vision teach {NLP} about efficient neural networks?
\newblock In: Proceedings of SustaiNLP: Workshop on Simple and Efficient Natural Language Processing, pp. 124--135. Association for Computational Linguistics, Online (2020).
\newblock \doi{10.18653/v1/2020.sustainlp-1.17}.
\newblock \urlprefix\url{https://aclanthology.org/2020.sustainlp-1.17}

\bibitem{jawahar2019does}
Jawahar, G., Sagot, B., Seddah, D.: What does {BERT} learn about the structure of language?
\newblock In: Proceedings of the 57th Annual Meeting of the Association for Computational Linguistics, pp. 3651--3657. Association for Computational Linguistics, Florence, Italy (2019).
\newblock \doi{10.18653/v1/P19-1356}.
\newblock \urlprefix\url{https://aclanthology.org/P19-1356}

\bibitem{jiao2020tinybert}
Jiao, X., Yin, Y., Shang, L., Jiang, X., Chen, X., Li, L., Wang, F., Liu, Q.: {T}iny{BERT}: Distilling {BERT} for natural language understanding.
\newblock In: Findings of the Association for Computational Linguistics: EMNLP 2020, pp. 4163--4174. Association for Computational Linguistics, Online (2020).
\newblock \doi{10.18653/v1/2020.findings-emnlp.372}.
\newblock \urlprefix\url{https://aclanthology.org/2020.findings-emnlp.372}

\bibitem{jin2021lifelong}
Jin, X., Zhang, D., Zhu, H., Xiao, W., Li, S.W., Wei, X., Arnold, A., Ren, X.: Lifelong pretraining: Continually adapting language models to emerging corpora.
\newblock In: Proceedings of BigScience Episode {\#}5 -- Workshop on Challenges {\&} Perspectives in Creating Large Language Models, pp. 1--16. Association for Computational Linguistics, virtual+Dublin (2022).
\newblock \doi{10.18653/v1/2022.bigscience-1.1}.
\newblock \urlprefix\url{https://aclanthology.org/2022.bigscience-1.1}

\bibitem{kim2021bert}
Kim, S., Gholami, A., Yao, Z., Mahoney, M.W., Keutzer, K.: {I-BERT:} integer-only {BERT} quantization.
\newblock In: M.~Meila, T.~Zhang (eds.) Proceedings of the 38th International Conference on Machine Learning, {ICML} 2021, 18-24 July 2021, Virtual Event, \emph{Proceedings of Machine Learning Research}, vol. 139, pp. 5506--5518. {PMLR} (2021).
\newblock \urlprefix\url{http://proceedings.mlr.press/v139/kim21d.html}

\bibitem{kumar2017data}
Kumar, A., Boehm, M., Yang, J.: Data management in machine learning: Challenges, techniques, and systems.
\newblock In: S.~Salihoglu, W.~Zhou, R.~Chirkova, J.~Yang, D.~Suciu (eds.) Proceedings of the 2017 {ACM} International Conference on Management of Data, {SIGMOD} Conference 2017, Chicago, IL, USA, May 14-19, 2017, pp. 1717--1722. {ACM} (2017).
\newblock \doi{10.1145/3035918.3054775}.
\newblock \urlprefix\url{https://doi.org/10.1145/3035918.3054775}

\bibitem{albert}
Lan, Z., Chen, M., Goodman, S., Gimpel, K., Sharma, P., Soricut, R.: {ALBERT:} {A} lite {BERT} for self-supervised learning of language representations.
\newblock In: 8th International Conference on Learning Representations, {ICLR} 2020, Addis Ababa, Ethiopia, April 26-30, 2020. OpenReview.net (2020).
\newblock \urlprefix\url{https://openreview.net/forum?id=H1eA7AEtvS}

\bibitem{lee2020biobert}
Lee, J., Yoon, W., Kim, S., Kim, D., Kim, S., So, C.H., Kang, J.: Biobert: a pre-trained biomedical language representation model for biomedical text mining.
\newblock Bioinformatics \textbf{36}(4), 1234--1240 (2020)

\bibitem{lee2018pretzel}
Lee, Y., Scolari, A., Chun, B.G., Santambrogio, M.D., Weimer, M., Interlandi, M.: $\{$PRETZEL$\}$: Opening the black box of machine learning prediction serving systems.
\newblock In: 13th USENIX Symposium on Operating Systems Design and Implementation (OSDI 18), pp. 611--626 (2018)

\bibitem{lester2021power}
Lester, B., Al-Rfou, R., Constant, N.: The power of scale for parameter-efficient prompt tuning.
\newblock In: Proceedings of the 2021 Conference on Empirical Methods in Natural Language Processing, pp. 3045--3059. Association for Computational Linguistics, Online and Punta Cana, Dominican Republic (2021).
\newblock \doi{10.18653/v1/2021.emnlp-main.243}.
\newblock \urlprefix\url{https://aclanthology.org/2021.emnlp-main.243}

\bibitem{bertemd}
Li, J., Liu, X., Zhao, H., Xu, R., Yang, M., Jin, Y.: {BERT}-{EMD}: Many-to-many layer mapping for {BERT} compression with earth mover{'}s distance.
\newblock In: Proceedings of the 2020 Conference on Empirical Methods in Natural Language Processing (EMNLP), pp. 3009--3018. Association for Computational Linguistics, Online (2020).
\newblock \doi{10.18653/v1/2020.emnlp-main.242}.
\newblock \urlprefix\url{https://aclanthology.org/2020.emnlp-main.242}

\bibitem{li2021intermittent}
Li, L., Nakandala, S., Kumar, A.: Intermittent human-in-the-loop model selection using cerebro: a demonstration.
\newblock Proceedings of the VLDB Endowment \textbf{14}(12) (2021)

\bibitem{li2021towards}
Li, S., Kumar, A.: Towards an optimized group by abstraction for large-scale machine learning.
\newblock Proceedings of the VLDB Endowment \textbf{14}(11), 2327--2340 (2021)

\bibitem{li2021prefix}
Li, X.L., Liang, P.: Prefix-tuning: Optimizing continuous prompts for generation.
\newblock In: Proceedings of the 59th Annual Meeting of the Association for Computational Linguistics and the 11th International Joint Conference on Natural Language Processing (Volume 1: Long Papers), pp. 4582--4597. Association for Computational Linguistics, Online (2021).
\newblock \doi{10.18653/v1/2021.acl-long.353}.
\newblock \urlprefix\url{https://aclanthology.org/2021.acl-long.353}

\bibitem{li2020automating}
Li, Y., Han, Z., Zhang, Q., Li, Z., Tan, H.: Automating cloud deployment for deep learning inference of real-time online services.
\newblock In: IEEE INFOCOM 2020-IEEE Conference on Computer Communications, pp. 1668--1677. IEEE (2020)

\bibitem{roberta}
Liu, Y., Ott, M., Goyal, N., Du, J., Joshi, M., Chen, D., Levy, O., Lewis, M., Zettlemoyer, L., Stoyanov, V.: Roberta: A robustly optimized bert pretraining approach.
\newblock arXiv preprint  (2019)

\bibitem{liu2022veltair}
Liu, Z., Leng, J., Zhang, Z., Chen, Q., Li, C., Guo, M.: Veltair: towards high-performance multi-tenant deep learning services via adaptive compilation and scheduling.
\newblock In: Proceedings of the 27th ACM International Conference on Architectural Support for Programming Languages and Operating Systems, pp. 388--401 (2022)

\bibitem{lo-wang-2020-s2orc}
Lo, K., Wang, L.L., Neumann, M., Kinney, R., Weld, D.: {S}2{ORC}: The semantic scholar open research corpus.
\newblock In: Proceedings of the 58th Annual Meeting of the Association for Computational Linguistics, pp. 4969--4983. Association for Computational Linguistics, Online (2020).
\newblock \doi{10.18653/v1/2020.acl-main.447}.
\newblock \urlprefix\url{https://aclanthology.org/2020.acl-main.447}

\bibitem{nakandala2020cerebro}
Nakandala, S., Zhang, Y., Kumar, A.: Cerebro: A data system for optimized deep learning model selection.
\newblock Proceedings of the VLDB Endowment \textbf{13}(12), 2159--2173 (2020)

\bibitem{Narayan2018DontGM}
Narayan, S., Cohen, S.B., Lapata, M.: Don't give me the details, just the summary! topic-aware convolutional neural networks for extreme summarization.
\newblock ArXiv \textbf{abs/1808.08745} (2018)

\bibitem{olston2017tensorflow}
Olston, C., Fiedel, N., Gorovoy, K., Harmsen, J., Lao, L., Li, F., Rajashekhar, V., Ramesh, S., Soyke, J.: Tensorflow-serving: Flexible, high-performance ml serving.
\newblock ArXiv preprint \textbf{abs/1712.06139} (2017).
\newblock \urlprefix\url{https://arxiv.org/abs/1712.06139}

\bibitem{paster2023openwebmath}
Paster, K., Santos, M.D., Azerbayev, Z., Ba, J.: Openwebmath: An open dataset of high-quality mathematical web text (2023).
\newblock \urlprefix\url{https://arxiv.org/abs/2310.06786}

\bibitem{qin2021bibert}
Qin, H., Ding, Y., Zhang, M., Qinghua, Y., Liu, A., Dang, Q., Liu, Z., Liu, X.: Bibert: Accurate fully binarized bert.
\newblock In: International Conference on Learning Representations (2021)

\bibitem{gpt}
Radford, A., Narasimhan, K., Salimans, T., Sutskever, I.: Improving language understanding by generative pre-training.
\newblock Tech. rep., OpenAI (2018)

\bibitem{reed1993pruning}
Reed, R.: Pruning algorithms-a survey.
\newblock IEEE transactions on Neural Networks \textbf{4}(5), 740--747 (1993)

\bibitem{rogers2020primer}
Rogers, A., Kovaleva, O., Rumshisky, A.: A primer in {BERT}ology: What we know about how {BERT} works.
\newblock Transactions of the Association for Computational Linguistics \textbf{8}, 842--866 (2020).
\newblock \doi{10.1162/tacl_a_00349}.
\newblock \urlprefix\url{https://aclanthology.org/2020.tacl-1.54}

\bibitem{ruckle2020adapterdrop}
R{\"u}ckl{\'e}, A., Geigle, G., Glockner, M., Beck, T., Pfeiffer, J., Reimers, N., Gurevych, I.: {AdapterDrop}: {O}n the efficiency of adapters in transformers.
\newblock In: Proceedings of the 2021 Conference on Empirical Methods in Natural Language Processing, pp. 7930--7946. Association for Computational Linguistics, Online and Punta Cana, Dominican Republic (2021).
\newblock \doi{10.18653/v1/2021.emnlp-main.626}.
\newblock \urlprefix\url{https://aclanthology.org/2021.emnlp-main.626}

\bibitem{sanh2019distilbert}
Sanh, V., Debut, L., Chaumond, J., Wolf, T.: Distilbert, a distilled version of bert: smaller, faster, cheaper and lighter.
\newblock ArXiv preprint \textbf{abs/1910.01108} (2019).
\newblock \urlprefix\url{https://arxiv.org/abs/1910.01108}

\bibitem{schelter2015challenges}
Schelter, S., Biessmann, F., Januschowski, T., Salinas, D., Seufert, S., Szarvas, G.: On challenges in machine learning model management.
\newblock IEEE Data Engineering Bulletin  (2015).
\newblock \urlprefix\url{https://www.amazon.science/publications/on-challenges-in-machine-learning-model-management}

\bibitem{sheng2023s}
Sheng, Y., Cao, S., Li, D., Hooper, C., Lee, N., Yang, S., Chou, C., Zhu, B., Zheng, L., Keutzer, K., et~al.: S-lora: Serving thousands of concurrent lora adapters.
\newblock arXiv preprint arXiv:2311.03285  (2023)

\bibitem{stephenson2021on}
Stephenson, C., suchismita padhy, Ganesh, A., Hui, Y., Tang, H., Chung, S.: On the geometry of generalization and memorization in deep neural networks.
\newblock In: International Conference on Learning Representations (2021).
\newblock \urlprefix\url{https://openreview.net/forum?id=V8jrrnwGbuc}

\bibitem{sun2019bertpkd}
Sun, S., Cheng, Y., Gan, Z., Liu, J.: Patient knowledge distillation for {BERT} model compression.
\newblock In: Proceedings of the 2019 Conference on Empirical Methods in Natural Language Processing and the 9th International Joint Conference on Natural Language Processing (EMNLP-IJCNLP), pp. 4323--4332. Association for Computational Linguistics, Hong Kong, China (2019).
\newblock \doi{10.18653/v1/D19-1441}.
\newblock \urlprefix\url{https://aclanthology.org/D19-1441}

\bibitem{mobilebert}
Sun, Z., Yu, H., Song, X., Liu, R., Yang, Y., Zhou, D.: {M}obile{BERT}: a compact task-agnostic {BERT} for resource-limited devices.
\newblock In: Proceedings of the 58th Annual Meeting of the Association for Computational Linguistics, pp. 2158--2170. Association for Computational Linguistics, Online (2020).
\newblock \doi{10.18653/v1/2020.acl-main.195}.
\newblock \urlprefix\url{https://aclanthology.org/2020.acl-main.195}

\bibitem{touvron2023llama}
Touvron, H., Martin, L., Stone, K., \textit{et al.}: Llama 2: Open foundation and fine-tuned chat models (2023)

\bibitem{vartak2016modeldb}
Vartak, M.: {MODELDB:} {A} system for machine learning model management.
\newblock In: {CIDR} 2017, 8th Biennial Conference on Innovative Data Systems Research, Chaminade, CA, USA, January 8-11, 2017, Online Proceedings. www.cidrdb.org (2017).
\newblock \urlprefix\url{http://cidrdb.org/cidr2017/gongshow/abstracts/cidr2017\_112.pdf}

\bibitem{self_attn}
Vaswani, A., Shazeer, N., Parmar, N., Uszkoreit, J., Jones, L., Gomez, A.N., Kaiser, L., Polosukhin, I.: Attention is all you need.
\newblock In: I.~Guyon, U.~von Luxburg, S.~Bengio, H.M. Wallach, R.~Fergus, S.V.N. Vishwanathan, R.~Garnett (eds.) Advances in Neural Information Processing Systems 30: Annual Conference on Neural Information Processing Systems 2017, December 4-9, 2017, Long Beach, CA, {USA}, pp. 5998--6008 (2017)

\bibitem{vaswani2017attention}
Vaswani, A., Shazeer, N., Parmar, N., Uszkoreit, J., Jones, L., Gomez, A.N., Kaiser, L., Polosukhin, I.: Attention is all you need.
\newblock In: I.~Guyon, U.~von Luxburg, S.~Bengio, H.M. Wallach, R.~Fergus, S.V.N. Vishwanathan, R.~Garnett (eds.) Advances in Neural Information Processing Systems 30: Annual Conference on Neural Information Processing Systems 2017, December 4-9, 2017, Long Beach, CA, {USA}, pp. 5998--6008 (2017).
\newblock \urlprefix\url{https://proceedings.neurips.cc/paper/2017/hash/3f5ee243547dee91fbd053c1c4a845aa-Abstract.html}

\bibitem{wang2022skipbert}
Wang, J., Chen, K., Chen, G., Shou, L., McAuley, J.: {S}kip{BERT}: Efficient inference with shallow layer skipping.
\newblock In: Proceedings of the 60th Annual Meeting of the Association for Computational Linguistics (Volume 1: Long Papers), pp. 7287--7301. Association for Computational Linguistics, Dublin, Ireland (2022).
\newblock \doi{10.18653/v1/2022.acl-long.503}.
\newblock \urlprefix\url{https://aclanthology.org/2022.acl-long.503}

\bibitem{wang2023smile}
Wang, J., Chen, K., Shou, L., Jiang, D., Chen, G.: Smile: A cost-effective system for serving massive pretrained language models in the cloud (demo).
\newblock In: Proceedings of the 2023 International Conference on Management of Data (2023)

\bibitem{wang2020minilm}
Wang, W., Wei, F., Dong, L., Bao, H., Yang, N., Zhou, M.: Minilm: Deep self-attention distillation for task-agnostic compression of pre-trained transformers.
\newblock In: H.~Larochelle, M.~Ranzato, R.~Hadsell, M.~Balcan, H.~Lin (eds.) Advances in Neural Information Processing Systems 33: Annual Conference on Neural Information Processing Systems 2020, NeurIPS 2020, December 6-12, 2020, virtual (2020).
\newblock \urlprefix\url{https://proceedings.neurips.cc/paper/2020/hash/3f5ee243547dee91fbd053c1c4a845aa-Abstract.html}

\bibitem{XiaoTCHL24}
Xiao, G., Tian, Y., Chen, B., Han, S., Lewis, M.: Efficient streaming language models with attention sinks.
\newblock In: The Twelfth International Conference on Learning Representations, {ICLR} 2024, Vienna, Austria, May 7-11, 2024. OpenReview.net (2024).
\newblock \urlprefix\url{https://openreview.net/forum?id=NG7sS51zVF}

\bibitem{xue2016managing}
Xue, J., Birke, R., Chen, L.Y., Smirni, E.: Managing data center tickets: Prediction and active sizing.
\newblock In: 2016 46th Annual IEEE/IFIP International Conference on Dependable Systems and Networks (DSN), pp. 335--346. IEEE (2016)

\bibitem{xlnet}
Yang, Z., Dai, Z., Yang, Y., Carbonell, J.G., Salakhutdinov, R., Le, Q.V.: Xlnet: Generalized autoregressive pretraining for language understanding.
\newblock In: H.M. Wallach, H.~Larochelle, A.~Beygelzimer, F.~d'Alch{\'{e}}{-}Buc, E.B. Fox, R.~Garnett (eds.) Advances in Neural Information Processing Systems 32: Annual Conference on Neural Information Processing Systems 2019, NeurIPS 2019, December 8-14, 2019, Vancouver, BC, Canada, pp. 5754--5764 (2019).
\newblock \urlprefix\url{https://proceedings.neurips.cc/paper/2019/hash/dc6a7e655d7e5840e66733e9ee67cc69-Abstract.html}

\bibitem{zafrir2019q8bert}
Zafrir, O., Boudoukh, G., Izsak, P., Wasserblat, M.: Q8bert: Quantized 8bit bert.
\newblock In: 2019 Fifth Workshop on Energy Efficient Machine Learning and Cognitive Computing-NeurIPS Edition (EMC2-NIPS), pp. 36--39. IEEE (2019)

\bibitem{zhang2019mark}
Zhang, C., Yu, M., Wang, W., Yan, F.: $\{$MArk$\}$: Exploiting cloud services for $\{$Cost-Effective$\}$,$\{$SLO-Aware$\}$ machine learning inference serving.
\newblock In: 2019 USENIX Annual Technical Conference (USENIX ATC 19), pp. 1049--1062 (2019)

\bibitem{Zhang00S0CM24}
Zhang, J., Wang, J., Li, H., Shou, L., Chen, K., Chen, G., Mehrotra, S.: Draft{\&} verify: Lossless large language model acceleration via self-speculative decoding.
\newblock In: Proceedings of the 62nd Annual Meeting of the Association for Computational Linguistics (Volume 1: Long Papers), {ACL} 2024, Bangkok, Thailand, August 11-16, 2024 (2024)

\bibitem{zhang2022opt}
Zhang, S., Roller, S., Goyal, N., Artetxe, M., Chen, M., Chen, S., Dewan, C., Diab, M., Li, X., Lin, X.V., et~al.: Opt: Open pre-trained transformer language models.
\newblock ArXiv preprint \textbf{abs/2205.01068} (2022).
\newblock \urlprefix\url{https://arxiv.org/abs/2205.01068}

\bibitem{zhou2022pets}
Zhou, Z., Wei, X., Zhang, J., Sun, G.: $\{$PetS$\}$: A unified framework for $\{$Parameter-Efficient$\}$ transformers serving.
\newblock In: 2022 USENIX Annual Technical Conference (USENIX ATC 22), pp. 489--504 (2022)

\end{thebibliography}

\end{sloppypar}
\end{document}